%% file: main.tex
\documentclass[10pt,twocolumn,letterpaper]{article}

\usepackage[pagenumbers]{cvpr} %

\usepackage{graphicx}
\usepackage{amsmath}
\usepackage{amssymb}
\usepackage{booktabs}
\usepackage{multirow}
\usepackage{adjustbox}

\usepackage[utf8]{inputenc} %
\usepackage[T1]{fontenc}    %
\usepackage{url}            %
\usepackage{booktabs}       %
\usepackage{amsfonts}       %
\usepackage{nicefrac}       %
\usepackage{microtype}      %
\usepackage{xcolor}         %
\usepackage{xspace}
\usepackage{mathtools}
\usepackage{bbm}

\definecolor{cvprblue}{rgb}{0.21,0.49,0.74}
\usepackage[pagebackref,breaklinks,colorlinks,citecolor=cvprblue]{hyperref}
\usepackage{colortbl}
\definecolor{lightgray}{rgb}{0.95,0.95,0.95} %

\usepackage[capitalize]{cleveref}
\crefname{section}{Sec.}{Secs.}
\Crefname{section}{Section}{Sections}
\Crefname{table}{Table}{Tables}
\crefname{table}{Tab.}{Tabs.}

\makeatletter
\DeclareRobustCommand\onedot{\futurelet\@let@token\@onedot}
\def\@onedot{\ifx\@let@token.\else.\null\fi\xspace}

\def\eg{\emph{e.g}\onedot}

\makeatother

\usepackage{pifont}
\newcommand{\cmark}{\ding{51}}%
\newcommand{\xmark}{\ding{55}}%

\usepackage{soul}

\def\paperID{4590} %
\def\confName{CVPR}
\def\confYear{2024}

\begin{document}

\title{MA-LMM: Memory-Augmented Large Multimodal Model \\ for Long-Term Video Understanding}

\newcommand{\system}{MA-LMM\xspace}

\author{Bo He$^{1, 2}$\footnotemark[1]\quad Hengduo Li$^{2}$\quad Young Kyun Jang$^{2}$\quad Menglin Jia$^{2}$\quad Xuefei Cao$^{2}$\quad \\ 
Ashish Shah$^{2}$\quad Abhinav Shrivastava$^{1}$\quad Ser-Nam Lim$^{3}$\\[0.5em]
$^{1}$University of Maryland, College Park \quad\quad $^{2}$Meta \quad\quad $^{3}$University of Central Florida\\
{\tt\small ~\url{https://boheumd.github.io/MA-LMM/}}
}

\maketitle

\renewcommand{\thefootnote}{\fnsymbol{footnote}}
\footnotetext[1]{Work done during Bo's internship at Meta.}

\input{sec/0_abstract}

\input{sec/1_introduction}
\input{sec/2_related}
\input{sec/3_method}

\input{sec/4_results}

\input{sec/5_conclusion}

\medskip
\noindent\textbf{Acknowledgements.}  This project was partially funded by NSF CAREER Award (\#2238769) to AS.

{
    \small
    \bibliography{main}
}

\clearpage
\input{appendix}

\end{document}

%% file: sec/0_abstract.tex
\begin{abstract}
With the success of large language models (LLMs), integrating the vision model into LLMs to build vision-language foundation models has gained much more interest recently. However, existing LLM-based large multimodal models (\eg, Video-LLaMA, VideoChat) can only take in a limited number of frames for short video understanding. In this study, we mainly focus on designing an efficient and effective model for long-term video understanding. Instead of trying to process more frames simultaneously like most existing work, we propose to process videos in an online manner and store past video information in a memory bank. This allows our model to reference historical video content for long-term analysis without exceeding LLMs' context length constraints or GPU memory limits. Our memory bank can be seamlessly integrated into current multimodal LLMs in an off-the-shelf manner. We conduct extensive experiments on various video understanding tasks, such as long-video understanding, video question answering, and video captioning, and our model can achieve state-of-the-art performances across multiple datasets. 
\end{abstract}

%% file: sec/1_introduction.tex
\section{Introduction}
\label{sec:intro}

Large language models (LLMs) have gained significant popularity in the natural language processing field.
By pre-training on large-scaled textual data, LLMs (\eg GPT~\cite{radford2018improving,radford2019language,brown2020language,chatgpt}, LLaMA~\cite{touvron2023llama1,touvron2023llama2}) have demonstrated remarkable abilities to perform both generative and discriminative tasks with a unified framework. 
Recently, there has been a growing interest in utilizing LLMs on multimodal tasks. By integrating LLMs with visual encoders, they can take images and videos as input and show incredible capabilities in various visual understanding tasks, such as captioning, question answering~\cite{li2023blip,liu2023visual,instructblip,zhu2023minigpt,ye2023mplug,zhang2023video,2023videochat}, classification, detection, and segmentation~\cite{chen2023videollm,wang2023visionllm,chen2023minigpt,wang2022omnivl,wang2022efficient,wang2023omnitracker,wang2024omnivid}.

\begin{figure}[t]
\centering
    \begin{subfigure}[b]{\linewidth}
        \vspace{-0.15in}
        \adjincludegraphics[width=\linewidth, trim={{0.0\width} {0.02\height} {0.0\width} {0.0\height}},clip]{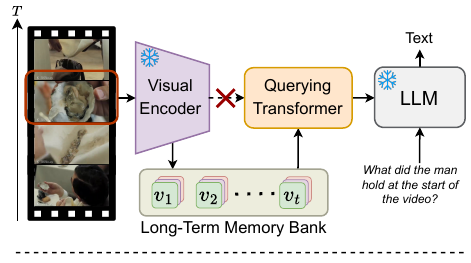}
        \vspace{-0.3in}
    \end{subfigure}
    \vskip\baselineskip
    \begin{subfigure}[b]{\linewidth}
        \adjincludegraphics[width=\linewidth, trim={{0.0\width} {0.01\height} {0.0\width} {0.02\height}},clip]{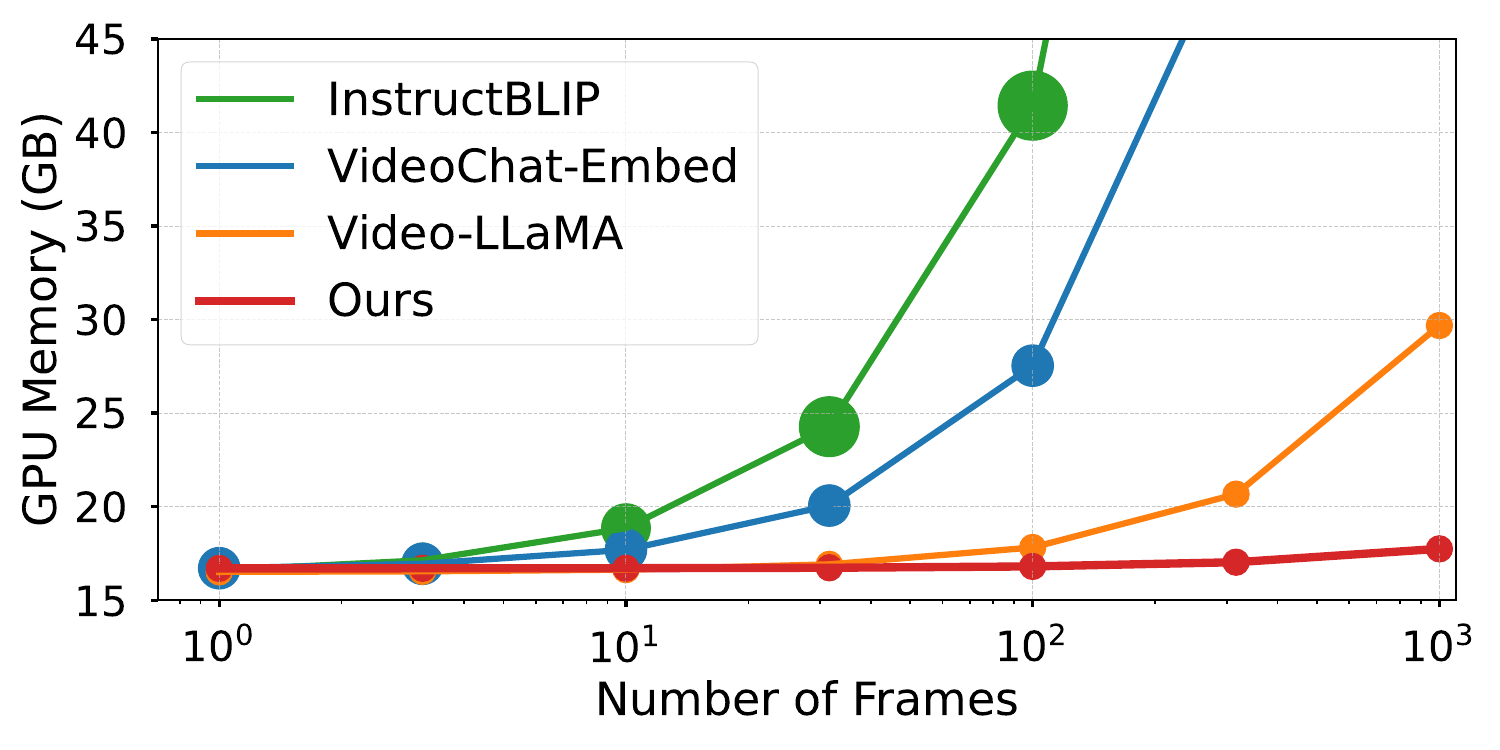}
        \vspace{-0.32in}
    \end{subfigure}
    \caption{(a) We propose the long-term memory bank to auto-regressively store and accumulate past video information, different from previous methods directly feeding the visual encoder's outputs into the querying transformer. (b) GPU memory and token number v.s. video frame length of multimodal methods and \system during inference. Circle sizes represent the number of text tokens.}
    \label{fig:teaser}
    \vspace{-0.3in}
\end{figure}

To handle video inputs, some prior large multimodal models~\cite{li2023blip,instructblip} directly feed the concatenated query embeddings of each frame along the temporal axis into LLMs. However, the inherent context length limitation of LLMs and GPU memory consumption restrict the number of video frames that can be processed. For example, LLaMA has a context length limitation of 2048 while large multimodal models like LLaVA~\cite{liu2023visual} and BLIP-2~\cite{li2023blip,instructblip} take in 256 and 32 tokens per image respectively.
Therefore, this design is not practical and feasible when video duration is much longer (\eg movies and TV shows).
To address these issues, a naive solution is to apply average pooling along the temporal axis like Video-ChatGPT~\cite{maaz2023video}, but this leads to inferior performances as it lacks explicit temporal modeling.
An alternative method involves adding a video modeling component to capture temporal dynamics, as seen in Video-LLaMA~\cite{zhang2023video}, which employs an extra video querying transformer (Q-Former) to obtain video-level representation.
However, this design adds model complexities, increases the training parameters, and is not suitable for online video analysis.

With these in mind, we introduce a \textbf{M}emory-\textbf{A}ugmented \textbf{L}arge \textbf{M}ultimodal \textbf{M}odel (\system), aiming for efficient and effective long-term video modeling. 
\system adopts a structure similar to existing large multimodal models~\cite{li2023blip,instructblip,zhang2023video}, which comprise a visual encoder to extract visual features, a querying transformer to align the visual and text embedding spaces, and a large language model. 
As illustrated in Figure~\ref{fig:teaser}\textcolor{red}{(a)}, as opposed to directly feeding visual encoder outputs to the querying transformer, we opt for an online processing approach that takes video frames sequentially and stores the video features in the proposed long-term memory bank.
This strategy of sequentially processing video frames and leveraging a memory bank significantly reduces the GPU memory footprint for long video sequences. 
It also effectively addresses the constraints posed by the limited context length in LLMs as demonstrated in Figure~\ref{fig:teaser}\textcolor{red}{(b)}.
Our design provides a solution for long-term video understanding with large multimodal models with great advantages over prior approaches~\cite{li2023blip,instructblip,zhang2023video,maaz2023video,2023videochat} which consume huge GPU memory and require a large number of input text tokens.

The core contribution of our approach is the introduction of a long-term memory bank that captures and aggregates historical video information. 
Specifically, the memory bank aggregates past video features in an auto-regressive manner, which can be referenced during subsequent video sequence processing.
Also, our memory bank is designed to be compatible with the Q-Former, where it acts as the key and value in the attention operation for long-term temporal modeling. 
As a result, it can be seamlessly integrated into existing large multimodal models in an off-the-shelf manner to enable long-term video modeling ability.
To further enhance efficiency, we propose a memory bank compression method that maintains the length of the memory bank constant relative to the input video length. By selecting and averaging the most similar adjacent frame features, it can preserve all the temporal information while significantly reducing the temporal redundancies in long videos.

We summarize our main contributions as follows:
\begin{itemize}[leftmargin=0.3in,topsep=0.02in]
    \item We introduce a novel long-term memory bank design to enhance existing large multimodal models, equipping them with long-term video modeling capability. 
    \item Our model significantly reduces the GPU memory usage and addresses LLMs' context length limitations by processing video sequences in an online fashion.
    \item Our approach has achieved new state-of-the-art performances on various downstreaming video tasks, including long-term video understanding, video question answering, and video captioning.
\end{itemize}

%% file: sec/2_related.tex
\section{Related Work}
\label{sec:related}

\noindent\textbf{Image-language models.}
Inspired by the success of powerful language models~\cite{chatgpt,touvron2023llama1,touvron2023llama2,radford2018improving,radford2019language,brown2020language}, recent image-language models tend to incorporate pre-trained language models with image encoders to support the multimodal reasoning ability~\cite{alayrac2022flamingo,li2023blip,liu2023visual,instructblip,zhu2023minigpt}. 
Flamingo~\cite{alayrac2022flamingo} proposes to connect powerful pre-trained vision-only and language-only models and achieve state-of-the-art performance in few-shot learning tasks.
BLIP-2~\cite{li2023blip} introduces a lightweight querying transformer to bridge the modality gap between the frozen pre-trained image encoder and frozen LLMs. Despite having significantly fewer trainable parameters, it performs well on various multimodal tasks.
LLaVA~\cite{liu2023visual} employs a simple linear layer to project image features into the text embedding space and efficiently finetunes LLMs ~\cite{hu2021lora} for better performance.
Building upon BLIP-2, MiniGPT-4~\cite{zhu2023minigpt} collects a large-scale high-quality dataset of image-text pairs and achieves better language generation ability.  
VisionLLM~\cite{wang2023visionllm} leverages the reasoning and parsing capacities of LLMs, producing strong performance on multiple fine-grained object-level and coarse-grained reasoning tasks.

\vspace{0.05in}
\noindent\textbf{Video-language models.}
Previous image-language models such as Flamingo~\cite{alayrac2022flamingo} and BLIP-2~\cite{li2023blip,instructblip} can also support video inputs. 
They simply flattened the spatio-temporal features into 1D sequences and then fed them into the language models for video inputs.
However, these approaches can not effectively capture the temporal dynamics of videos.
Based on this motivation, Video-LLaMA~\cite{zhang2023video} enhances BLIP-2 structure by adding an additional video querying transformer to explicitly model the temporal relationship.
Similarly, building on LLaVA~\cite{liu2023visual}, Video-ChatGPT~\cite{maaz2023video} simply average pools the frame-level features across spatial and temporal dimensions to generate video-level representation.
VideoChat~\cite{2023videochat} utilizes perception models to generate action and object annotations, which are then forwarded to LLMs for further reasoning. 
Despite the advancements, these models are primarily designed for short videos.
Inspired by the Token Merging~\cite{bolya2022token} which averages highly similar tokens to reduce the computation cost, we propose an extension of this idea to video data, specifically along the temporal axis. 
This extension aims to mitigate the challenges posed by extensive token numbers and computational cost associated with processing long video inputs. Several concurrent works~\cite{ren2023testa,song2023moviechat,jin2023chat} have also explored similar strategies of merging akin tokens for video inputs. 
Please refer to the supplementary material for more detailed discussions.

\begin{figure*}[t!]
    \centering
    \vspace{-0.15in}
    \includegraphics[width=0.99\textwidth]{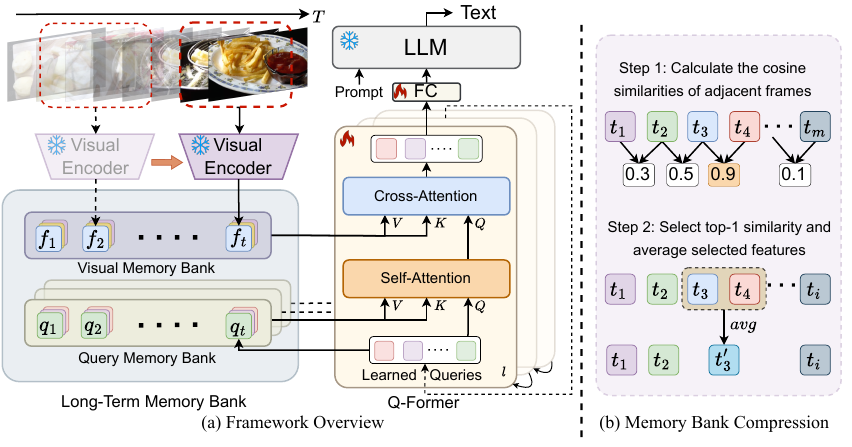}
    \vspace{-0.15in}
    \caption{(a) Framework overview. \system auto-regressively processes video frames in an online manner. Two long-term memory banks are designed to store the raw visual features and learned queries at each timestep, which are used for future reference. The Q-Former is composed of several cascaded blocks, indexed by $l$. LLM outputs text for various video understanding downstream tasks. The snowflake icon indicates components with fixed parameters, while the flame icon denotes parts of the model that are fine-tuned. (b) Illustration of the memory bank compression technique, which is applied to maintain the length of the memory bank constant.}
    \label{fig:overview}
    \vspace{-0.15in}
\end{figure*}

\vspace{0.05in}
\noindent\textbf{Long-term video models.}
Long-term video understanding methods focus on capturing long-range patterns in long videos, which typically exceed 30 seconds. 
To mitigate the computational demands of processing long videos, a prevalent approach involves using pre-extracted features, sidestepping the need for joint training of backbone architectures~\cite{donahue2015long,yue2015beyond,girdhar2017actionvlad,hussein2019timeception,wu2021towards}.
Alternatively, some research works aim to devise sparse video sampling methods~\cite{korbar2019scsampler,wu2019adaframe}, reducing the number of input frames by only preserving salient video content.
Other works like Vis4mer~\cite{islam2022long} and S5~\cite{wang2023selective} leverage the streamlined transformer decoder structure of S4~\cite{gu2021efficiently} to enable long-range temporal modeling with linear computation complexity.
Inspired by the memory bank design~\cite{chen2020memory,lee2018memory,lee2021video,wu2022memvit}, we propose to integrate the long-term memory bank with large multimodal models to enable efficient and effective long-term temporal modeling capabilities.

%% file: sec/3_method.tex
\vspace{-0.05in}
\section{Method}
\vspace{-0.05in}

We introduce \system, a memory-augmented large multimodal model for long-term video understanding.
Instead of processing more frames simultaneously as most video understanding methods~\cite{feichtenhofer2016convolutional,hussein2019timeception,feichtenhofer2019slowfast,he2020gta,he2022asm,he2023align,bertasius2021space,fan2021multiscale,tong2022videomae}, we propose to auto-regressively process video frames in an online manner, which draws inspiration from the online processing fashion with long-term memory design presented in MeMViT~\cite{wu2022memvit}.
Figure~\ref{fig:overview}\textcolor{red}{(a)} illustrates the overview of our \system framework.
Following similar practices of large multimodal models~\cite{li2023blip,instructblip,liu2023visual,zhang2023video}, the overall model architecture can be divided into three parts: (1) visual feature extraction with a frozen visual encoder (Sec.~\ref{sec:visual_feature}), (2) long-term temporal modeling with a trainable querying transformer (Q-Former) to align the visual and text embedding spaces (Sec.~\ref{sec:temporal_modeling}), and (3) text decoding with a frozen large language model (Sec.~\ref{sec:llm_decoding}).

\subsection{Visual Feature Extraction}
\label{sec:visual_feature}
This design draws inspiration from the cognitive processes humans use to handle long-term visual information. Instead of concurrently processing extensive duration of signals, humans process them in a sequential manner, correlate current visual inputs with past memories for comprehension, and selectively retain salient information for subsequent reference~\cite{wu2022memvit}. 
Similarly, our \system processes video frames sequentially, dynamically associating new frame input with historical data stored in the long-term memory bank, ensuring that only discriminative information is conserved for later use. 
This selective retention facilitates a more sustainable and efficient approach to video understanding, which further allows the model to automatically support online video reasoning tasks.

Formally, given a sequence of $T$ video frames, we pass each video frame into a pre-trained visual encoder and obtain the visual features $V = [v_1, v_2, .., v_T], v_t \in \mathbb{R}^{P\times C}$, where $P$ is the number of patches for each frame and $C$ is the channel dimension for the extracted frame feature.
Then we inject temporal ordering information into the frame-level features by a position embedding layer ($PE$) as
\begin{align}
    f_t = v_t + PE(t), f_t \in \mathbb{R}^{P\times C}.
\end{align}

\subsection{Long-term Temporal Modeling}
\label{sec:temporal_modeling}
For aligning the visual embedding to the text embedding space, we use the same architecture as the Querying Transformer (Q-Former) in BLIP-2~\cite{li2023blip,instructblip}.
Q-Former takes in the learned queries $z \in \mathbb{R}^{N\times C}$ to capture video temporal information, where $N$ is the number of learned queries, and $C$ is the channel dimension.
In our experiments, Q-Former outputs 32 tokens for each image, which is more efficient than 256 tokens produced by LLaVA~\cite{liu2023visual}.
Each Q-Former block consists of two attention submodules:
(1) cross-attention layer, which interacts with the raw visual embedding extracted from the frozen visual encoder, and (2) self-attention layer, which models interactions within the input queries.
Different from the original Q-Former in BLIP-2 that only attends to the current frame's embedding, we design a long-term memory bank consisting of the visual memory bank and the query memory bank, which accumulates the past video information and augments the input to cross- and self-attention layers for effective long-term video understanding.

\vspace{0.05in}
\noindent\textbf{Visual Memory Bank.}
The visual memory bank stores the raw visual features of each frame extracted from the frozen visual encoder. 
Specifically, for the current time step $t$, the visual memory bank contains the concatenated list of past visual features $F_t = \texttt{Concat}[f_1, f_2, .., f_t], F_t \in \mathbb{R}^{tP\times C}$.
Given the input query $z_t$, the visual memory bank acts as the key and value as:
\vspace{-0.05in}
\begin{align}
    Q&=z_tW_Q,\  K=F_tW_K,\  V = F_tW_V. \label{eq:attention1}
\end{align}
Then we apply the cross-attention operation as:
\vspace{-0.05in}
\begin{align}
    O&=Attn(Q, K, V) = \text{Softmax}\left(\dfrac{QK^T}{\sqrt{C}}\right)V. \label{eq:attention2}
\end{align}
In this way, it can explicitly attend to past visual information through the cached visual memory bank with long-term context. 
Since all the cross-attention layers in the Q-Former attend to the same visual feature, there is only one visual memory bank that is shared across all the Q-Former blocks. 

\vspace{0.05in}
\noindent\textbf{Query Memory Bank.}
Different from the fixed visual memory bank which stores the raw and static visual features, the query memory bank accumulates input queries of each timestep, represented as $Z_t = \texttt{Concat}[z_1, z_2, .., z_t], Z_t \in \mathbb{R}^{tN\times C}$.
By storing these queries, we maintain a dynamic memory of the model's understanding and processing of each frame up to the current timestep via the Q-Former. 
The query memory bank also acts as key and value as:
\begin{align}
    Q&=z_tW_{Q},\  K=Z_tW_{K},\  V=Z_tW_{V}. \label{eQ_2:attention3}
\end{align}
similar to the Eq~\ref{eq:attention1}. Then we apply the same attention operation as Eq.~\ref{eq:attention2}.
At each time step, $z_t$ contains the learned important information specifically for each video till the current timestep $t$.
Different from the static visual memory bank, the input queries $z_t$ evolve through cascaded Q-Former blocks during the model training, capturing distinct video concepts and patterns at increasing levels of abstraction.
As a result, each self-attention layer has a unique query memory bank, where the contained input queries are updated during the training time.

\vspace{0.05in}
\noindent\textbf{Memory Bank Compression.}
Given that our model directly stores past video information in the memory banks, the GPU memory and computational cost increase linearly as the number of past video frames. This becomes particularly challenging for long videos, and thus it is essential to further compress the memory bank to a smaller size.
One conventional approach to managing temporal sequences involves employing a first-in-first-out queue. Here, features from the earliest time step are removed when the memory bank reaches a pre-defined limit, a strategy utilized in MeMViT~\cite{wu2022memvit}. However, it results in the loss of earlier historical information as new frames are added and old features are popped to maintain memory bank capacity.
Alternatively, MeMViT employs learnable pooling operators to compress the spatio-temporal size of stored feature in the memory bank, albeit at the cost of introducing additional trainable parameters.

Drawing inspiration from the effectiveness of token merging and pruning techniques showcased in works such as~\cite{bolya2022token,meng2022adavit,yin2022vit,rao2021dynamicvit}, we introduce a novel Memory Bank Compression (MBC) technique to exploit temporal redundancies inherent in videos.
Our proposed method aggregates and compresses video information over time by leveraging the similarity between adjacent features, thereby retaining early historical information. This approach effectively compresses repetitive information within the memory bank while preserving discriminative features. Notably, several concurrent works~\cite{ren2023testa,song2023moviechat,jin2023chat} have similarly embraced the token merging strategies to reduce video redundancies.

Same as MeMViT~\cite{wu2022memvit}, which applies feature compression at each iteration, our method applies the compression algorithm at each auto-regressive step if the current length of the memory bank exceeds the predefined threshold $M$.
Formally, given the visual memory bank containing a list of $[f_1, f_2, .., f_M], f_t \in \mathbb{R}^{P\times C}$, when a new frame feature $f_{M+1}$ comes in, we need to compress the memory bank by reducing the length by 1.
At each spatial location $i$, we first calculate the cosine similarity between all the temporally adjacent tokens as
\begin{equation}
    s^i_t = \cos(f^i_t, f^i_{t+1}), t \in [1, M], i \in [1, P].
\end{equation}
Then we select the highest similarity across time, which can be interpreted as the most temporally redundant features:
\begin{equation}
    k = \mathrm{argmax}_t(s^i_t).
\end{equation}
Next, we simply average the selected token features at all the spatial locations to reduce the memory bank length by 1:
\begin{equation}
    \hat{f}^i_k = (f^i_k + f^i_{k+1}) / 2. \label{eq:compression1}
\end{equation}
In this way, we can still preserve the most discriminative features while keeping the temporal ordering unchanged as depicted in Figure~\ref{fig:overview}\textcolor{red}{(b)}.
The same procedure is adopted to compress the query memory bank. 

\subsection{Text Decoding}
\label{sec:llm_decoding}
As we process video frames in an auto-regressive manner, the Q-Former output at the final timestep contains all historical information, which is then fed into the LLM.
Therefore, we can significantly reduce the number of input text tokens from $N * T$ to $N$, addressing the context length limitation of the current LLMs and substantially easing the GPU memory requirements.
During training, given a labeled dataset consisting of video and text pairs, our model is supervised with the standard cross entropy loss as:
\begin{equation}
    \mathcal{L} = -\frac{1}{S} \sum_{i=1}^{S} \log P(w_i | w_{<i}, V).
\end{equation}
in which $V$ represents the input video, and $w_i$ is the $i$-th ground-truth text token.
During training, we update the parameters of the Q-Former while keeping the weights of both the visual encoder and the language model frozen.

%% file: sec/4_results.tex
\section{Experiments}

\begin{table*}[t]
\centering
\vspace{-0.1in}
\begin{minipage}{.68\textwidth}
    \caption{Comparison with state-of-the-art methods on the LVU~\cite{wu2021towards} dataset. \textbf{Bold} and \underline{underline} represent the top-1 and top-2 results.}
    \vspace{-0.1in}
    \resizebox{0.988\linewidth}{!}{
    \renewcommand{\arraystretch}{1.15}
    \begin{tabular}{>{\kern-0.5\tabcolsep}l|ccccccc|c<{\kern-0.5\tabcolsep}}
        \toprule
        \multirow{2}{*}{\textbf{Model}} & \multicolumn{3}{c}{\textbf{Content}} & \multicolumn{4}{c|}{\textbf{Metadata}} & \multirow{2}{*}{Avg} \\
        \cmidrule(lr){2-4} \cmidrule(lr){5-8}
        &  Relation & Speak & Scene & Director & Genre & Writer & Year & \\
        \midrule
        Obj\_T4mer~\cite{wu2021towards} & 54.8 & 33.2 & 52.9 & 47.7 & 52.7 & 36.3 & 37.8 & 45.0 \\
        Performer~\cite{choromanski2020rethinking} & 50.0 & 38.8 & 60.5 & 58.9 & 49.5 & 48.2 & 41.3 & 49.6 \\
        Orthoformer~\cite{patrick2021keeping} & 50.0 & 38.3 & 66.3 & 55.1 & 55.8 & 47.0 & 43.4 & 50.8 \\
        VideoBERT~\cite{sun2019videobert} & 52.8 & 37.9 & 54.9 & 47.3 & 51.9 & 38.5 & 36.1 & 45.6 \\
        LST~\cite{islam2022long} & 52.5 & 37.3 & 62.8 & 56.1 & 52.7 & 42.3 & 39.2 & 49.0 \\
        VIS4mer~\cite{islam2022long} & 57.1 & 40.8 & 67.4 & 62.6 & 54.7 & 48.8 & 44.8 & 53.7 \\
        S5~\cite{wang2023selective} & \bf 67.1 & \underline{42.1} & \underline{73.5} & \underline{67.3} & \bf 65.4 & \underline{51.3} & \underline{48.0} & \underline{59.2} \\
        \midrule
        \rowcolor{lightgray} \bf Ours & \underline{58.2} & \bf 44.8 & \bf 80.3 & \bf 74.6 & \underline{61.0} & \bf 70.4 & \bf 51.9 & \bf 63.0 \\
        \bottomrule
    \end{tabular}
    \label{tab:lvu}
}
\end{minipage}
\hfill
\begin{minipage}{.3\textwidth}
    \caption{Comparison on the Breakfast~\cite{kuehne2014language} and COIN~\cite{tang2019coin} datasets. The top-1 accuracy is reported here.}
    \vspace{-0.1in}
    \resizebox{\linewidth}{!}{
    \renewcommand{\arraystretch}{1.25}
    \begin{tabular}{>{\kern-0.5\tabcolsep}l|cc<{\kern-0.5\tabcolsep}}
        \toprule
        \textbf{Model} & \textbf{Breakfast} & \textbf{COIN} \\
        \midrule
        TSN~\cite{wang2018temporal} & - & 73.4 \\
        VideoGraph~\cite{hussein2019videograph} & 69.5 & - \\
        Timeception~\cite{hussein2019timeception} & 71.3 & - \\
        GHRM~\cite{zhou2021graph} & 75.5 & - \\
        D-Sprv.~\cite{lin2022learning} & 89.9 & 90.0 \\
        ViS4mer~\cite{islam2022long} & 88.2 & 88.4 \\
        S5~\cite{wang2023selective} & \underline{90.7} & \underline{90.8} \\
        \midrule
        \rowcolor{lightgray} \bf Ours & \bf 93.0 & \bf 93.2 \\
        \bottomrule
    \end{tabular}
    }
    \label{tab:breakfast}
\end{minipage}
\end{table*}

\subsection{Tasks and Datasets}
To validate the effectiveness of the proposed \system, we mainly focus on the long-term video understanding task.
We also extend the evaluation to standard video understanding tasks (\eg, video question answering, video captioning) to further compare with existing multimodal methods.

\vspace{0.05in}
\noindent\textbf{Long-term Video Understanding.}
We conduct experiments on three widely used long-term video datasets including LVU~\cite{wu2021towards}, Breakfast~\cite{kuehne2014language}, and COIN~\cite{tang2019coin}. We report the top-1 classification accuracy as the evaluation metric.
The LVU dataset contains $\sim$30K videos extracted from $\sim$3K movies, with each video lasting 1 to 3 minutes. Given that current large multimodal models generally perform text generation and lack regression capability, we limit our experiments to seven classification tasks: relationship, speaking style, scene, director, genre, writer, and release year.
The Breakfast~\cite{kuehne2014language} dataset includes videos related to breakfast preparation, which consists of 1712 videos with an average length of around 2.7 minutes. 
COIN~\cite{tang2019coin} is a large-scale dataset for comprehensive instructional video analysis, which comprises 11827 instructional videos from YouTube, covering 180 distinct tasks in 12 domains related to daily life. The average length of a video is 2.36 minutes.

\vspace{0.05in}
\noindent\textbf{Video Question Answering.}
We conduct evaluation on three open-ended video question answering datasets including MSRVTT-QA~\cite{xu2017video}, MSVD-QA~\cite{xu2017video}, and ActivityNet-QA~\cite{yu2019activitynet}. ActivityNet-QA contains long videos with average durations of 2 minutes, while MSRVTT-QA and MSVD-QA consist of short videos with 10-15 seconds duration.

\vspace{0.05in}
\noindent\textbf{Video Captioning.}
We report the video captioning results of  METEOR~\cite{banerjee2005meteor} and CIDEr~\cite{vedantam2015cider} metrics on three popular datasets: MSRVTT~\cite{xu2016msr}, MSVD~\cite{chen2011collecting} and Youcook2~\cite{zhou2018towards}.

\vspace{0.05in}
\noindent\textbf{Online Action Prediction.}
We further evaluate the online prediction capability of our model by conducting experiments on the EpicKitchens-100~\cite{damen2020epic} dataset, which consists of 700 long videos of cooking activities with 100 total hours. It includes 97 verbs, 300 nouns, and 3807 action types. Following the same experimental setting in ~\cite{zhong2023anticipative}, we report the top-5 accuracy and recall results on the validation dataset.

\subsection{Implementation Details}
For the visual encoder, we adopt the pre-trained image encoder ViT-G/14~\cite{dosovitskiy2020image} from EVA-CLIP~\cite{fang2023eva}, it can be further changed to other clip-based video encoders.
We use the pre-trained Q-Former weights from InstructBLIP~\cite{instructblip} and adopt Vicuna-7B~\cite{vicuna2023} as the LLM. 
All the experiments are conducted on 4 A100 GPUs.
More details about training and evaluation are described in the supplementary material.

\begin{table*}[t]
\centering
\begin{minipage}{.45\textwidth}
    \centering
    \caption{Comparison with state-of-the-art methods on the video question answering task. Top-1 accuracy is reported.}
    \vspace{-0.1in}
    \resizebox{0.98\linewidth}{!}{
    \renewcommand{\arraystretch}{1.15}
    \begin{tabular}{>{\kern-0.5\tabcolsep}l|ccc<{\kern-0.5\tabcolsep}}
        \toprule
        \textbf{Model} & \textbf{MSRVTT} & \textbf{MSVD} & \textbf{ActivityNet} \\
        \midrule
        JustAsk~\cite{yang2021just} & 41.8 & 47.5 & 38.9 \\
        FrozenBiLM~\cite{yang2022zero} & 47.0 & 54.8 & 43.2 \\
        SINGULARITY~\cite{lei2022revealing} & 43.5 & -- & 44.1 \\
        VIOLETv2~\cite{fu2023empirical} & 44.5 & 54.7 & -- \\
        GiT~\cite{wang2022git} & 43.2 & 56.8 & -- \\
        mPLUG-2~\cite{xu2023mplug} & \underline{48.0} & 58.1 & -- \\
        UMT-L~\cite{li2023unmasked} & 47.1 & 55.2 & 47.9 \\
        VideoCoCa~\cite{yan2022video} & 46.3 & 56.9 & \bf 56.1 \\
        \midrule
        Video-LLaMA~\cite{zhang2023video} & 46.5 & \underline{58.3} & 45.5 \\
        \rowcolor{lightgray} \bf Ours & \bf 48.5 & \bf 60.6 & \underline{49.8} \\
        \bottomrule
    \end{tabular}
    }
    \label{tab:vqa}
\end{minipage}
\hfill
\begin{minipage}{.52\textwidth}
\centering
    \caption{Comparison with state-of-the-art methods on the video captioning task. METEOR (M) and CIDEr (C) results are reported.}
    \vspace{-0.1in}
    \resizebox{0.98\linewidth}{!}{
    \renewcommand{\arraystretch}{1.2}
    \begin{tabular}{>{\kern-0.5\tabcolsep}l|cc|cc|cc<{\kern-0.5\tabcolsep}}
        \toprule
        \multirow{2}{*}{\textbf{Model}} & \multicolumn{2}{c|}{\textbf{MSRVTT}} & \multicolumn{2}{c|}{\textbf{MSVD}} & \multicolumn{2}{c}{\textbf{YouCook2}} \\
        \cmidrule{2-7}
         & \textbf{M} & \textbf{C} & \textbf{M} & \textbf{C} & \textbf{M} & \textbf{C} \\
        \midrule
        UniVL~\cite{luo2020univl} & 28.2 & 49.9 & 29.3 & 52.8 & -- & 127.0 \\
        SwinBERT~\cite{lin2022swinbert} & 29.9 & 53.8 & 41.3 & 120.6 & 15.6 & 109.0 \\
        GIT~\cite{wang2022git} & 32.9 & 73.9 & \bf 51.1 & \bf 180.2 & \underline{17.3} & \underline{129.8} \\
        mPLUG-2~\cite{xu2023mplug} & \bf 34.9 & \bf 80.3 & 48.4 & 165.8 & -- & -- \\
        VideoCoca~\cite{yan2022video} & -- & 73.2 & -- & -- & -- & 128.0 \\
        \midrule
        Video-LLaMA & 32.9 & 71.6 & 49.8 & 175.3 & 16.5 & 123.7 \\
        \rowcolor{lightgray} \bf Ours & \underline{33.4} & \underline{74.6} & \underline{51.0} & \underline{179.1} & \bf 17.6 & \bf 131.2 \\
        \bottomrule
    \end{tabular}
    }
    \label{tab:caption}
\end{minipage}
\vspace{-0.1in}
\end{table*}

\subsection{Main Results}
\noindent\textbf{Long-term Video Understanding.}
We compare \system with previous state-of-the-art (SOTA) methods on the LVU benchmark~\cite{wu2021towards} in Table~\ref{tab:lvu}.
Notably, \system outperforms existing long-term video models (S5~\cite{wang2023selective}, ViS4mer~\cite{islam2022long}, VideoBERT~\cite{sun2019videobert}, and Object Transformer~\cite{wu2021towards}) in both content understanding and metadata prediction tasks. This results in significant improvement in most tasks, enhancing the average top-1 accuracy by 3.8\% compared to the S5~\cite{wang2023selective} model.
Unlike previous video-based models which process all video frames simultaneously in an offline manner and predict probabilities for each class, our \system processes video frames in an online fashion and directly outputs the text label for each class type.

We also evaluate our \system on the Breakfast~\cite{kuehne2014language} and COIN~\cite{tang2019coin} datasets that pose a challenge for the long-term video activity classification task. 
We show the results in Table~\ref{tab:breakfast}. Our method improves upon the previous best method, S5\cite{wang2023selective}, by 2.3\% and 2.4\% respectively on the top-1 accuracy metric. This result further proves the superior long-term video understanding capability of our approach.

\vspace{0.05in}
\noindent\textbf{Video Question Answering.}
To compare with existing multimodal video understanding methods, we conduct experiments on the open-ended video question answering datasets in Table~\ref{tab:vqa} to demonstrate the generalization ability of our model. Given that these are mostly short videos, it is expected that our memory bank will be less effective.
Interestingly, we observe that our \system achieves new state-of-the-art performances on the MSRVTT and MSVD datasets while falling short of VideoCoCa's performance on the ActivityNet dataset.
On the latter, it is not surprising, since VideoCoCa~\cite{yan2022video} leverages large-scale video-text datasets for pre-training (\eg, HowTo100M~\cite{miech2019howto100m} and VideoCC3M~\cite{nagrani2022learning}) while our \system uses model weights only pre-trained on the image-text datasets.

Notably, our \system significantly outperforms the recent LLM-based model Video-LLaMA~\cite{zhang2023video} on all three datasets.
Video-LLaMA concatenates all the query embeddings from the frozen image Q-Former and trains an additional video Q-Former from scratch to model temporal dependencies, consuming too much GPU memory to be feasible for long video inputs.
In contrast, our \system simply fine-tunes the weights from the pre-trained image Q-Former without introducing an additional video Q-Former, yet is able to effectively capture temporal relationships by virtue of the long-term memory bank.
This result strongly justifies the superiority of our design on the general video question answering task, and reveals that even a few frames and queries captured in the memory banks can have significant beneficial effects.

\begin{table}[t]
\centering
    \caption{Action anticipation results on EpicKitchens-100.}
    \label{tab:action_prediction}
    \vspace{-0.1in}
    \resizebox{\linewidth}{!}{
    \renewcommand{\arraystretch}{1.2}
        \begin{tabular}{>{\kern-0.5\tabcolsep}l|ccc|ccc<{\kern-0.5\tabcolsep}}
            \toprule
            \multirow{2}{*}{\textbf{Model}} & \multicolumn{3}{c|}{\textbf{Accuracy@Top-5}} & \multicolumn{3}{c}{\textbf{Recall@Top-5}} \\
            \cmidrule(lr){2-4} \cmidrule(lr){5-7}
            & Verb & Noun & Act. & Verb & Noun & Act. \\
            \midrule
            Video-LLaMA & 73.9 & 47.5 & 29.7 & \bf 26.3 & 27.3 & 11.7 \\
            \rowcolor{lightgray} \bf Ours & \bf 74.5 & \bf 50.7 & \bf 32.7 & 25.9 & \bf 29.9 & \bf 12.2 \\
            \bottomrule
        \end{tabular}
    }
\vspace{-0.2in}
\end{table}

\begin{table*}[t]
\centering
\vspace{-0.1in}
\begin{minipage}{.45\textwidth}
    \centering
    \caption{Contribution of visual and query memory banks.}
    \vspace{-0.1in}
    \label{tab:components_ablation}
    \resizebox{0.95\linewidth}{!}{
        \renewcommand{\arraystretch}{1.2}
        \begin{tabular}{>{\kern-0.5\tabcolsep}cc|ccc<{\kern-0.5\tabcolsep}}
            \toprule
            \bf Visual & \bf Query & \bf LVU & \bf Breakfast & \bf COIN \\
            \midrule
             \xmark & \xmark & 48.3 & 74.6 & 72.3 \\
             \cmark & \xmark & 61.5 & 91.8 & 92.4 \\
             \xmark & \cmark & 58.0 & 81.4 & 88.5 \\
            \rowcolor{lightgray} \cmark & \cmark & \bf 63.0 & \bf 93.0 & \bf 93.2 \\
            \bottomrule
        \end{tabular}
    }
    \vfill
    \vspace{0.15in}
    \centering
    \caption{Contribution of the long-term memory bank (MB) under off-the-shelf evaluation without training.}
    \vspace{-0.1in}
    \label{tab:zero_shot}
    \resizebox{0.95\linewidth}{!}{
        \renewcommand{\arraystretch}{1.2}
        \begin{tabular}{>{\kern-0.5\tabcolsep}c|cccc<{\kern-0.5\tabcolsep}}
            \toprule
            \textbf{MB} & \textbf{MSRVTT} & \textbf{MSVD} & \textbf{ActivityNet} & \textbf{LVU} \\
            \midrule
            \xmark  & 19.5 & 38.8 & 29.9 & 23.6 \\
            \rowcolor{lightgray} \cmark & \bf 20.3  & \bf 40.0  & \bf 37.2 & \bf 32.8  \\
            \bottomrule
        \end{tabular}
    }
\end{minipage}
\hfill
\begin{minipage}{.53\textwidth}
    \caption{Ablation of different temporal modeling methods.}
    \label{tab:temmporal_ablation}
    \vspace{-0.1in}
    \resizebox{\linewidth}{!}{
        \renewcommand{\arraystretch}{1.2}
        \begin{tabular}{>{\kern-0.5\tabcolsep}c|c|c|c|ccc<{\kern-0.5\tabcolsep}}
            \toprule
            \bf Method & \bf \#Frame & \bf \#Token & \bf GPU & \bf LVU & \bf Breakfast & \bf COIN \\
            \midrule
             Concat & 60 & 1920 & 49.2 & 62.6 & 90.4 & 93.0 \\
             Avg Pool & 100 & 32 & 21.2 & 57.6 & 80.6 & 87.6 \\
             ToMe & 100 & 200 & 22.2 & 61.5 & 91.3 & 91.5 \\
             \midrule
             FIFO & 100 & 32 & 19.1 & 61.3 & 88.5 & 90.4 \\
             \rowcolor{lightgray} MBC & 100 & 32 & 19.1 & \bf 63.0 & \bf 93.0 & \bf 93.2 \\
            \bottomrule
        \end{tabular}
    }
    \vfill
    \centering
    \vspace{0.2in}
    \caption{The comparison of using different LLMs.}
    \vspace{-0.1in}
    \label{tab:llm}
    \resizebox{0.95\linewidth}{!}{
        \renewcommand{\arraystretch}{1.2}
        \begin{tabular}{>{\kern-0.5\tabcolsep}c|cccc<{\kern-0.5\tabcolsep}}
            \toprule
            \textbf{LLM} & \textbf{MSRVTT} & \textbf{MSVD} & \textbf{ActivityNet} & \textbf{LVU} \\
            \midrule
            FlanT5-XL & 46.5 & 57.6 & 48.2 & 62.0 \\
            \rowcolor{lightgray} Vicuna-7B & \bf 48.5 & \bf 60.6 & \bf 49.8 & \bf 63.0 \\
            \bottomrule
        \end{tabular}
    }
\end{minipage}
\vspace{-0.15in}
\end{table*}

\vspace{0.05in}
\noindent\textbf{Video Captioning.}
To further evaluate the capabilities of our \system in generating free-form text, we conduct experiments on the standard video captioning datasets including MSRVTT~\cite{xu2016msr}, MSVD~\cite{chen2011collecting} and YouCook2~\cite{zhou2018towards} in Table~\ref{tab:caption}.
Although these datasets only consist of videos with short duration and our model is initially pre-trained merely on image-text dataset pairs, our \system exhibits outstanding performances across all the metrics. It consistently ranks among the top-2 positions compared to current leading methods. Remarkably, our results also surpass the recent Video-LLaMA~\cite{zhang2023video} on these datasets, highlighting the significant improvements our model offers in both video captioning and question-answering tasks.

\vspace{0.05in}
\noindent\textbf{Online Action Prediction.}
Since our model can naturally support the online video understanding task, we compare our \system with Video-LLaMA on the EpicKitchens-100~\cite{damen2020epic} dataset to investigate the online action prediction capability. 
In Table~\ref{tab:action_prediction}, our \system outperforms Video-LLaMA, achieving more accurate results in both top-5 accuracy and recall measures. This highlights our model's superior capacity to anticipate actions in an online manner, showcasing its effectiveness for applications that require real-time analytical capabilities.

\subsection{Ablation Studies}
\noindent\textbf{Contribution of each component.}
To further investigate the contribution of the visual memory bank and query memory bank, we conduct ablation studies in Table~\ref{tab:components_ablation}.
Initially, we observe that without any memory bank module, the performances across all three datasets are notably worse, due to the lack of temporal context.
The introduction of either memory bank results in substantial improvements, confirming their roles in enhancing the model's ability to understand temporal sequences.
We also find that the visual memory bank achieves better performance than the query memory bank.
We hypothesize that the explicit method of storing historical raw video features in the visual memory bank is more effective than the query memory bank which implicitly captures video information through the input learned queries. 
And two memory banks are complementary to each other. 
When incorporating two memory banks together, our approach can boost the final performance by 14.7\%, 18.4\%, and 20.9\% on the LVU, Breakfast, and COIN, respectively.

\vspace{0.05in}
\noindent\textbf{Long-term temporal modeling ablation.}
We compare different temporal modeling approaches in Table~\ref{tab:temmporal_ablation}.
In our setup, the Q-Former outputs 32 text tokens per frame.
The most straightforward approach for temporal feature integration is either concatenating or averaging frame-level features.
However, they resulted in inferior performances.
Notably, concatenation requires a significantly higher number of text tokens and computational cost compared to other variants, which also introduces higher GPU memory consumption since they need to takes in all the video frames simultaneously.
In addition, we conduct experiments using ToMe~\cite{bolya2022token} to reduce the number of text tokens per frame from 32 to 2. 
However, without our auto-regressive strategy, it still requires 200 text tokens for 100-frame input. 
The second part of this table presents the performances of different memory bank compression approaches.
The first-in-first-out (FIFO) technique removes the oldest features to main the length of the memory bank fixed, while the memory bank compression (MBC) strategy  merges temporally consecutive features with the highest similarity, effectively reducing the most redundant information while keeping the temporal ordering unchanged.
With this design that theoretically keeps all historical information, MBC outperforms FIFO by 1.7\%, 4.5\%, and 2.8\% accuracy across three datasets.
This experimental result validates the superior efficiency and effectiveness of our approach in modeling long-term temporal information.

\begin{figure}[t]
\centering
    \adjincludegraphics[width=\linewidth, trim={{0.01\width} {0.02\height} {0.01\width} {0.01\height}},clip]{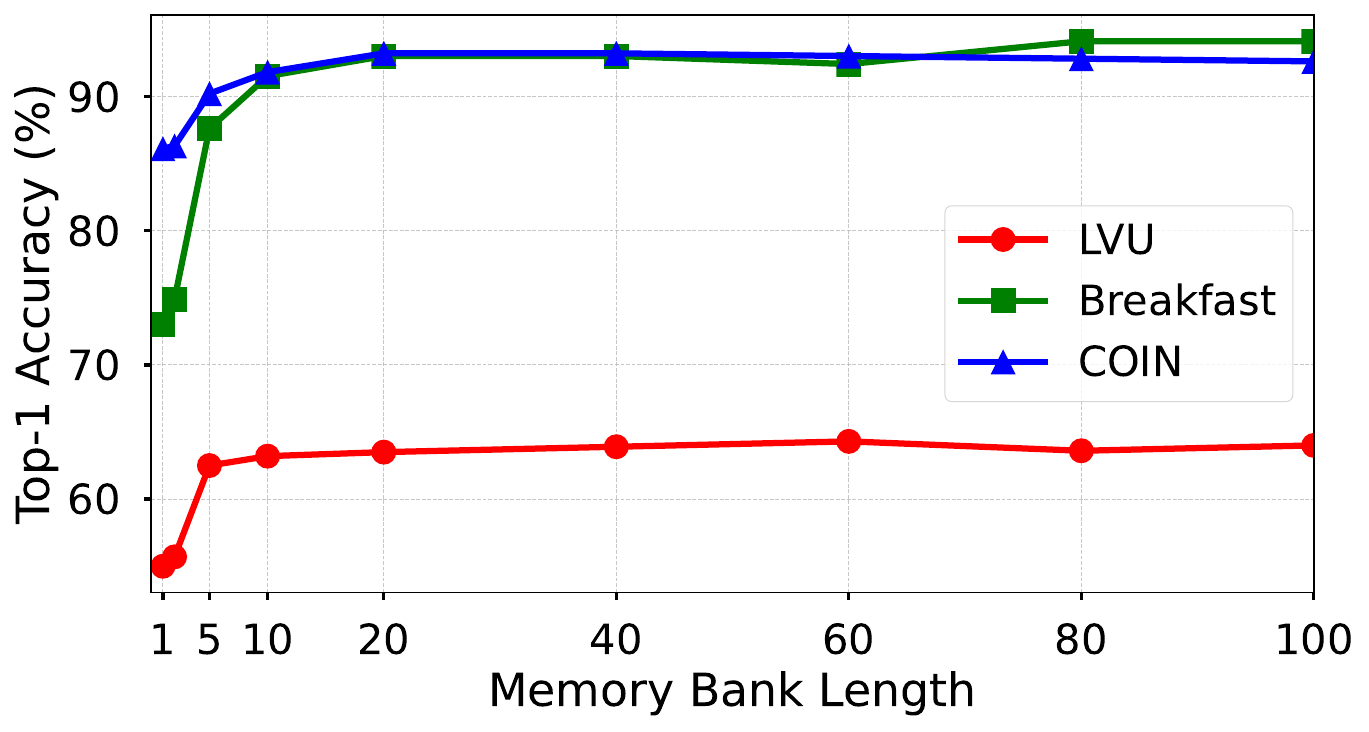}
    \vspace{-0.25in}
    \caption{Impact of different memory bank lengths.}
    \vspace{-0.3in}
    \label{fig:memory_bank}
\end{figure}

\begin{figure*}[t]
\vspace{-0.15in}
\centering
    \adjincludegraphics[width=\linewidth, trim={{0.0\width} {0.10\height} {0.0\width} {0.05\height}},clip]{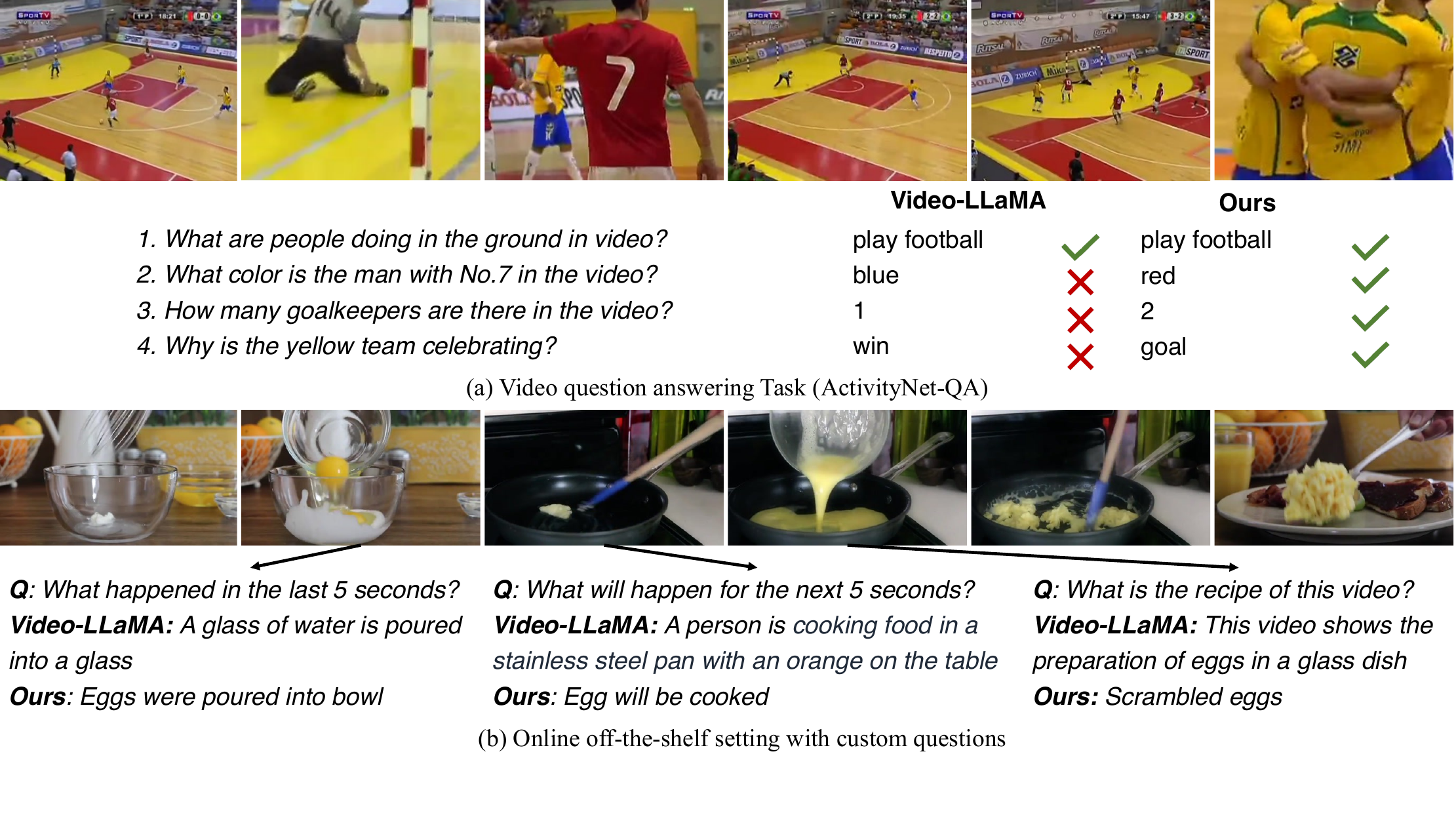}
    \vspace{-0.25in}
    \caption{Visualization results on the video question answering task and the online off-the-shelf setting.}
\vspace{-0.1in}
\label{fig:visualization}
\end{figure*}

\begin{figure*}[t]
\centering
    \adjincludegraphics[width=\linewidth, trim={{0.0\width} {0.0\height} {0.0\width} {0.0\height}},clip]{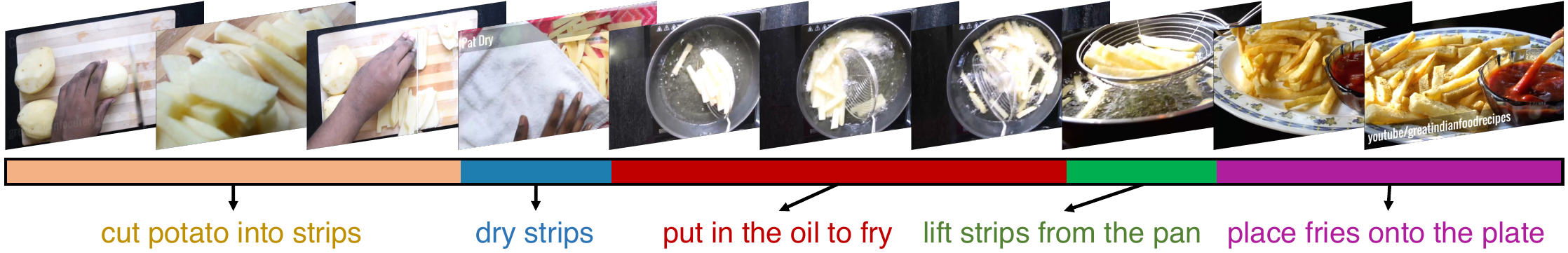}
    \vspace{-0.25in}
    \caption{Visualization of the compressed visual memory bank.}
\vspace{-0.2in}
\label{fig:visualization_compression}
\end{figure*}

\vspace{0.05in}
\noindent\textbf{Off-the-shelf evaluation.}
A key advantage of \system is that our long-term memory bank can be inserted into existing large multimodal models in an off-the-shelf manner, thereby endowing them with effective temporal modeling capabilities without retraining.
As presented in Table~\ref{tab:zero_shot}, \system can consistently boost the final performance when incorporating the long-term memory bank to the baseline method~\cite{instructblip}.
Particularly, on long-term video datasets like ActivityNet and LVU, \system can largely improve the results by 7.3\% and 9.2\%.
This highlights the robustness of long-term memory banks in temporal modeling under the off-the-shelf setting.

\vspace{0.05in}
\noindent\textbf{Different language model architectures.}
Our \system can utilize different language model architectures including but not limited to encoder-decoder models and decoder-only models. We experimented with two popular models FlanT5-XL~\cite{chung2022scaling} and Vicuna-7B~\cite{vicuna2023}, and show the results in Table~\ref{tab:llm} that the Vicuna-7B marginally outperforms the FlanT5-XL on these video tasks.

\vspace{0.05in}
\noindent\textbf{Memory bank length ablation.}
In Figure~\ref{fig:memory_bank}, we conduct experiments to evaluate the effect of varying the memory bank length.
Given an input of 100 video frames, the top-1 accuracy first increases as the feature bank length becomes larger. 
This rise can be attributed to the augmented storage capacity of the memory bank, which can preserve more historical data and consequently boost the final performance.
However, we observe that performances begin to saturate when the memory bank length is around 10 to 20. 
This supports our hypothesis that there are prevalent temporal redundancies in long videos, and we can significantly reduce the frame length without sacrificing the performance.

\subsection{Visualization}
In Figure~\ref{fig:visualization}, we provide a comprehensive visual comparison between \system and Video-LLaMA~\cite{zhang2023video}.
In the video question answering task, \system exhibits superior memorization and recognition capabilities. 
Specifically, it can accurately memorize historical information and recognize fine-grained information, such as the color of the man with No.7, and precisely count the number of goalkeepers who appeared in the video.
With the auto-regressive design, our model supports online reasoning directly. This capability is further exemplified in our experiments on off-the-shelf evaluations using custom questions. \system can correctly anticipate the next step of the video (\textit{"egg will be cooked"}) and predict the correct recipe (\textit{"scrambled egg"}).
More visualization examples are shown in the supplementary material.

Figure~\ref{fig:visualization_compression} provides a visualization of the compressed visual memory bank. We set the memory bank length to 5 for this illustration. The compressed visual memory bank appears to group consecutive frames with similar visual content. For instance, in the presented video, the video frames are effectively grouped into five clusters, each capturing a distinct yet semantically consistent activity, which is similar to the effect of temporal segmentation.

%% file: sec/5_conclusion.tex
\section{Conclusion}
\vspace{-0.05in}
In this paper, we introduce a long-term memory bank designed to augment current large multimodal models, equipping them with the capabilities to effectively and efficiently model long-term video sequences.
Our approach processes video frames sequentially and stores historical data in the memory bank, addressing LLMs' context length limitation and GPU memory constraints posed by the long video inputs. 
Our long-term memory bank is a plug-and-play module that can be easily integrated into existing large multimodal models in an off-the-shelf manner.
Experiments on various tasks have demonstrated the superior advantages of our method.
We believe our \system offers valuable insights for future research in the long-term video understanding area.

%% file: appendix.tex
\appendix
\section*{Appendix}

We present additional ablation experiments in Section~\ref{sec:addtional_results} and further qualitative results for the video captioning task in Section~\ref{sec:qualitative_more}.
Next, in Section~\ref{sec:relations}, we discuss the relations to concurrent works~\cite{ren2023testa,song2023moviechat,jin2023chat} in details.
And in Sec.~\ref{sec:experiment_details}, we show more dataset-specific implementation details and hyper-parameter settings.
Finally, we address some limitations and outline directions for future research in Section~\ref{sec:limitation}.

\section{Additional Experiments}
\label{sec:addtional_results}

\vspace{0.1in}
\noindent\textbf{Memory bank compression at different spatial levels.}
In Table~\ref{tab:finegrain}, we show comparison results of compressing the memory bank at different spatial levels (frame-level vs. token-level) on the LVU~\cite{wu2021towards}, Breakfast~\cite{kuehne2014language} and COIN~\cite{tang2019coin} datasets. 
For the frame-level compression, we calculate the cosine similarity between adjacent frame features and average the frame-level features with the highest similarity. 
For the token-level compression, the cosine similarity is calculated between tokens at the same spatial location across the entire temporal axis, given that each frame-level feature contains multiple tokens at different spatial locations.
The results indicate that token-level compression consistently surpasses frame-level compression in performance.
Particularly, on the Breakfast dataset, the token-level surpasses the frame-level by 6.5\% in top-1 accuracy. 
This superiority can be attributed to the importance of recognizing the object type of breakfast in videos.
And token-level compression can help preserve much more fine-grained spatial information and details.

\begin{figure}[t]
\centering
    \adjincludegraphics[width=0.95\linewidth, trim={{0.01\width} {0.05\height} {0.01\width} {0.01\height}},clip]{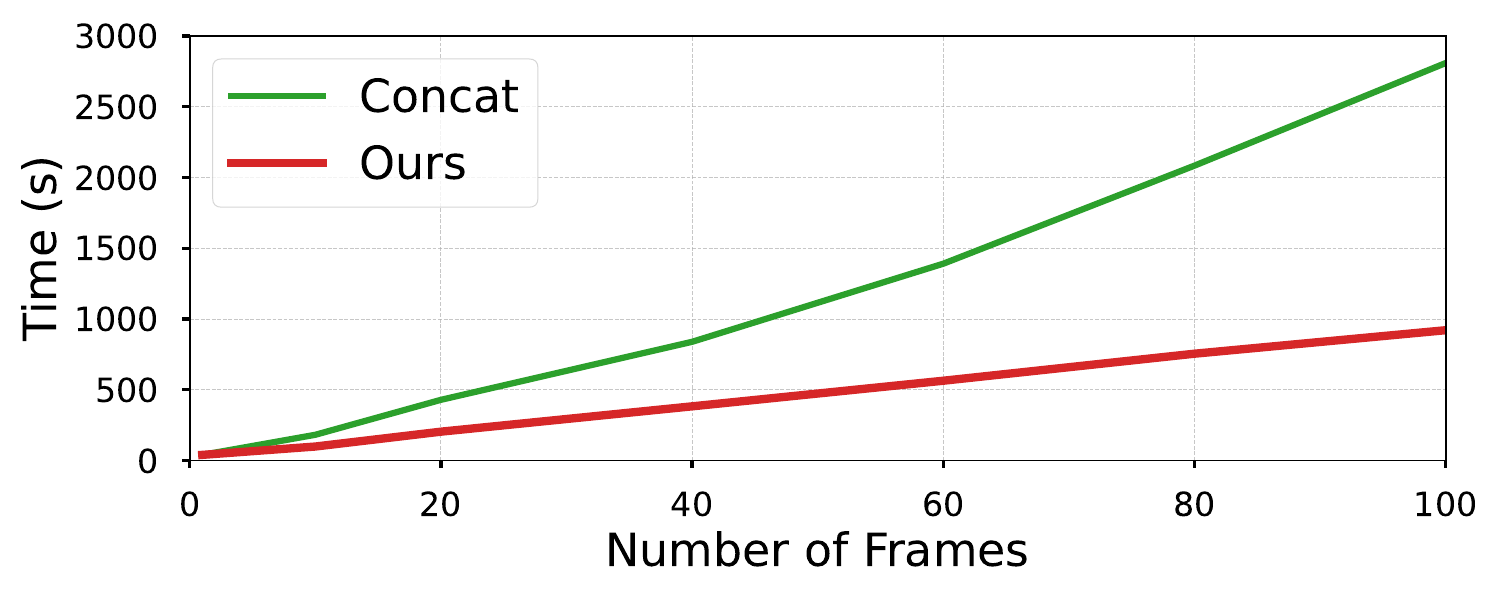}
    \vspace{-0.1in}
    \caption{Inference time vs. input frame length.}
    \vspace{-0.15in}
    \label{fig:time}
\end{figure}

\vspace{0.1in}
\noindent\textbf{Inference time v.s. video frame lengths.}
In Figure~\ref{fig:time}, the inference time of \system increases linearly with respect to the frame lengths, due to its auto-regressive design of processing video frames sequentially.
In contrast, directly concatenating frame-level features takes much longer time and higher GPU memory consumption, since it needs to process all video frames simultaneously.

\section{More Qualitative Results}
\label{sec:qualitative_more}

\begin{figure*}[t]
\centering
    \adjincludegraphics[width=\linewidth, trim={{0.0\width} {0.0\height} {0.0\width} {0.0\height}},clip]{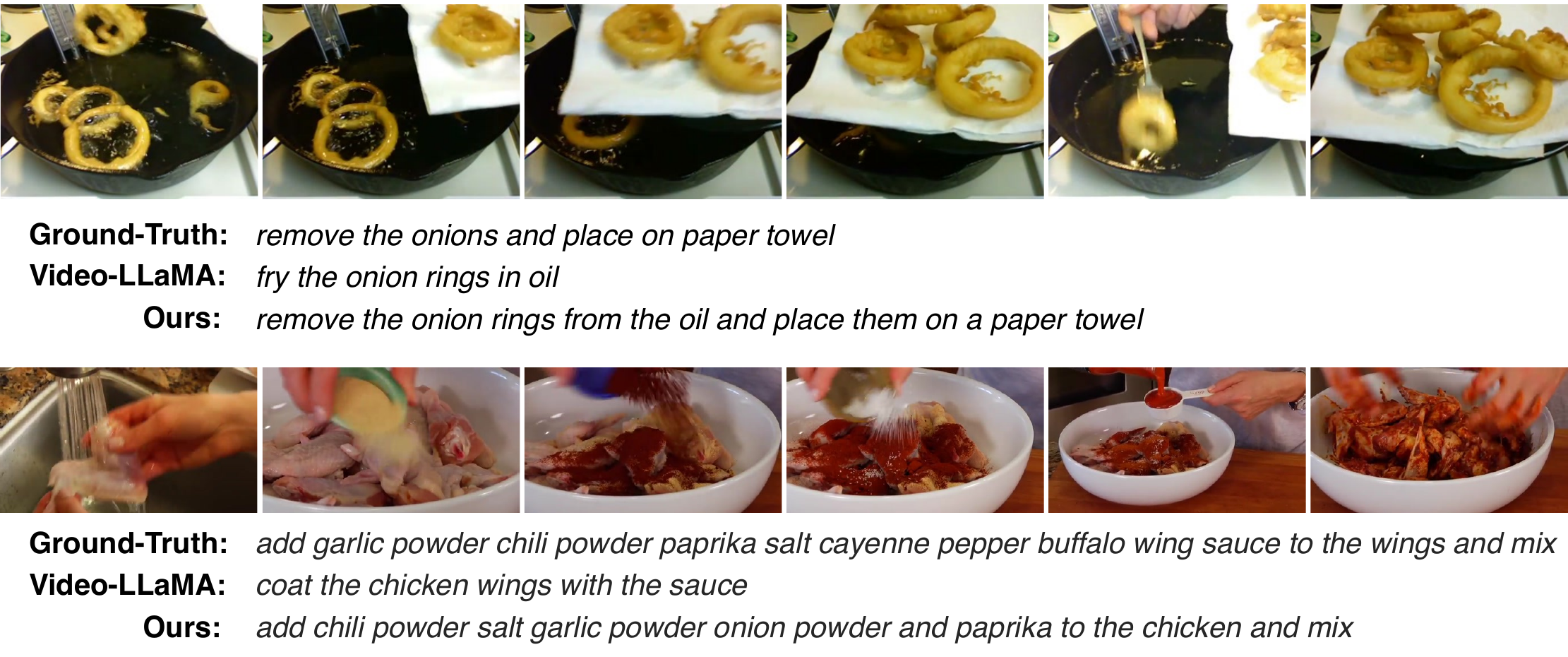}
    \vspace{-0.2in}
    \caption{Visualization results on the video captioning task.}
\vspace{-0.1in}
\label{fig:visualization_cap_supp}
\end{figure*}

Our model's enhanced capabilities in video captioning are further showcased through additional visualization results in Figure~\ref{fig:visualization_cap_supp}. 
Here, our \system significantly outperforms Video-LLaMA~\cite{zhang2023video} in generating detailed and accurate sentence descriptions. 
For instance, in the first video, our model precisely describes the action as "remove the onion rings and place them on the paper towel," capturing the entire action steps, while Video-LLaMA's description lacks this completeness, notably missing the crucial action of removing the onion rings. 
In the second video example, our model distinguishes itself by accurately identifying subtle details such as specific ingredients: chili powder, salt, and garlic powder, which Video-LLaMA overlooks. 
This highlights the enhanced capability of our \system in recognizing and describing fine-grained details.

\begin{table}[t]
    \centering
    \caption{Memory bank compression at different spatial levels.}
    \vspace{-0.05in}
    \renewcommand{\arraystretch}{1.2}
    \begin{tabular}{c|ccc}
        \toprule
        \bf Spatial Level & \bf LVU & \bf Breakfast & \bf COIN \\
        \midrule
         Frame-level & 61.8 & 86.5 & 91.1 \\
         \rowcolor{lightgray} Token-level & \bf 63.0 & \bf 93.0 & \bf 93.2 \\
        \bottomrule
    \end{tabular}
    \label{tab:finegrain}
\end{table}

\section{Relations to Concurrent Works}
\label{sec:relations}
In this section, we compare and discuss the relations between our \system with the concurrent works including TESTA~\cite{ren2023testa}, MovieChat~\cite{song2023moviechat} and Chat-UniVi~\cite{jin2023chat}. All of these methods focus on utilizing the idea of token merging~\cite{bolya2022token} to reduce video redundancies.

\vspace{0.05in}
\noindent\textbf{Temporal Modeling.}
Temporal modeling across these methodologies falls into three categories. 
Chat-UniVi~\cite{jin2023chat} directly feed visual tokens into large language models (LLMs) without explicit temporal modeling, utilizing LLMs' inherent sequence processing for video understanding. 
In contrast, TESTA~\cite{ren2023testa} and MovieChat~\cite{song2023moviechat} employ global self-attention; TESTA captures interactions along spatial and temporal dimensions, whereas MovieChat processes long videos in segments, compresses these into short-term memories, then concatenates and models global temporal interactions using a video Q-Former. 
Differently, our \system adopts causal self-attention, restricting each frame's feature access to prior video information only. Such a design naturally endows our \system with the capability to support online video applications in robotics, AR/VR, and video streaming.

\vspace{0.05in}
\noindent\textbf{Token Merging Application.}
Building on the token merging~\cite{bolya2022token} strategy, four methodologies have adopted and modified this approach to reduce video data redundancy. Each uses the core concept of merging similar tokens but differs in implementation. 
TESTA~\cite{ren2023testa} utilizes a cascaded module for spatial and temporal aggregation, progressively shortening video length and decreasing tokens per frame. 
In contrast, Chat-UniVi's~\cite{jin2023chat} modules operate in parallel, merging tokens across both dimensions before LLM reasoning. 
MovieChat~\cite{song2023moviechat} employs a selective strategy to merge similar adjacent frames, reducing the number of video frames. 
Similarly, our \system conducts token merging along the temporal dimension to condense video length but at a more fine-grained spatial level. It independently compresses visual and query tokens across different spatial areas, enhancing performance as evidenced in Table~\ref{tab:finegrain}.

\vspace{0.05in}
\noindent\textbf{Based Model.}
Both TESTA~\cite{ren2023testa} and Moviechat~\cite{song2023moviechat} are built upon the video-based multimodal model.
TESTA integrates TimeSFormer~\cite{bertasius2021space} as its video encoder, facilitating long-range video modeling. 
Meanwhile, MovieChat adopts the Video-LLaMA~\cite{zhang2023video} framework, combining an image Q-Former with a video Q-Former to effectively manage long-term temporal relationships. 
On the contrary, another group involves adapting image-based multimodal models for video understanding. Chat-UniVi~\cite{jin2023chat} leverages the LLaVA~\cite{liu2023visual} architecture, feeding concatenated visual tokens along the temporal axis into LLMs. 
Our \system builds on InstructBLIP~\cite{instructblip} as a plug-and-play module that significantly boosts long-term temporal modeling. Demonstrated in Table~\ref{tab:zero_shot}, our memory bank module greatly excels over InstructBLIP under the off-the-shelf setting without video-specific pre-training or introducing additional parameters.

\vspace{0.05in}
\noindent\textbf{Memory Bank Design.} 
The integration of memory banks to enhance long-term video understanding has been thoroughly explored~\cite{lee2018memory,wu2019long,gong2019memorizing,chen2020memory,wu2022memvit}. 
Building on these studies, MovieChat~\cite{song2023moviechat} and our \system both employ memory bank designs. MovieChat primarily uses memory banks to consolidate raw and static visual features. In contrast, our \system innovates with an additional query memory bank that captures dynamic memory, reflecting the evolving understanding of past video frames. The effectiveness of our query memory bank is evidenced in Table~\ref{tab:components_ablation}.

\section{Experiment Details}
\label{sec:experiment_details}
We build our \system on top of InstructBlip~\cite{instructblip}, following the codebase~\cite{li-etal-2023-lavis}.
We show the details of hyper-parameters in the following table for different tasks and datasets.
For all the experiments, we use a cosine learning rate decay.
Table~\ref{tab:hyper-lvu} shows the hyper-parameters for the long-term video understanding task. For the LVU dataset, we follow the same practice in~\cite{islam2022long,wang2023selective}, we sample 100 frames of 1 fps for each video clip. For the Breakfast~\cite{kuehne2014language} and COIN~\cite{tang2019coin}, we uniformly sample 100 frames from the whole video.
Table~\ref{tab:hyper-vqa} shows the hyper-parameters on the MSRVTT-QA~\cite{xu2017video}, MSVD-QA~\cite{xu2017video}, and ActivityNet-QA~\cite{yu2019activitynet} datasets for the video question answering task while Table~\ref{tab:hyper-caption} presents the hyperparameters on the MSRVTT~\cite{xu2016msr}, MSVD~\cite{chen2011collecting}, YouCook2~\cite{zhou2018towards} datasets for video captioning.

\section{Limitation and Future Work}
\label{sec:limitation}
Since our model takes in video frames in an online manner, leading to reduced GPU memory usage, but at the cost of increased video processing time. This trade-off becomes particularly noticeable with extremely long videos, where processing times can become significantly prolonged. 
To mitigate this issue, we suggest a hierarchical method to process extremely long-term video sequences.
This strategy involves dividing extensive videos into smaller segments and then processing each segment sequentially in an auto-regressive fashion as we present in the main paper. Then we can employ additional video modeling techniques to model inter-segment relationships. 
This method aims to strike a balance between memory efficiency and processing speed, making it a practical solution for long-term video understanding.

For the future work, there are several potential aspects to further enhance the model's capabilities. 
First, replacing the existing image-based visual encoder with a video or clip-based encoder can naturally enhance the model's ability to capture short-term video dynamics. This provides a better representation of the video's temporal dynamics.
Second, the model's overall performance in understanding videos can substantially benefit from the pre-training stage on large-scale video-text datasets. This approach is a common practice in existing research and has proven effective in enhancing generalization capabilities.
Finally, the flexibility inherent in our model's architecture allows for the incorporation of a more advanced LLM as the language decoder. This integration offers a clear opportunity for boosting the final performance, making our model more effective in interpreting and responding to complex video content.

\begin{table}[h]
\centering
    \caption{Hyperparameters of different datasets on the long-term video understanding task.}
    \vspace{-0.1in}
    \resizebox{\linewidth}{!}{
    \renewcommand{\arraystretch}{1.2}
    \begin{tabular}{@{}l|p{0.67in}p{0.67in}p{0.67in}@{}}
        \toprule
        \textbf{Dataset} & \textbf{LVU} & \textbf{Breakfast} & \textbf{COIN} \\ \midrule
        LLM & \multicolumn{3}{c}{Vicuna-7B} \\
        Epochs & \multicolumn{3}{c}{20} \\
        Learning rate    & \multicolumn{3}{c}{1e-4} \\
        Batch size   & \multicolumn{3}{c}{64}  \\
        AdamW \(\beta\)  & \multicolumn{3}{c}{(0.9, 0.999)} \\
        Weight decay & \multicolumn{3}{c}{0.05} \\
        Image resolution & \multicolumn{3}{c}{224} \\
        Beam size & \multicolumn{3}{c}{5} \\
        Frame length & \multicolumn{3}{c}{100} \\
        Memory bank length & \multicolumn{3}{c}{20} \\
        \multirow{4}{*}{Prompt} & ``What is the \{task\} of the movie?'' &  ``What type of breakfast is shown in the video?'' & ``What is the activity in the video?'' \\
        \bottomrule
    \end{tabular}
    }
    \label{tab:hyper-lvu}
\end{table}

\begin{table}[h]
\centering
    \vspace{-0.1in}
    \caption{Hyperparameters of different datasets on the video question answering task.}
    \vspace{-0.1in}
    \resizebox{\linewidth}{!}{
    \renewcommand{\arraystretch}{1.2}
    \begin{tabular}{@{}l|ccc@{}}
        \toprule
        \textbf{Dataset} & \textbf{MSRVTT} & \textbf{MSVD} & \textbf{ActivityNet} \\ \midrule
        LLM & \multicolumn{3}{c}{Vicuna-7B} \\
        Epochs & \multicolumn{3}{c}{5} \\
        Learning rate    & \multicolumn{3}{c}{1e-4} \\
        Batch size   & \multicolumn{3}{c}{128}  \\
        AdamW \(\beta\)  & \multicolumn{3}{c}{(0.9, 0.999)} \\
        Weight decay & \multicolumn{3}{c}{0.05} \\
        Image resolution & \multicolumn{3}{c}{224} \\
        Beam size & \multicolumn{3}{c}{5} \\
        Frame length &  & 20 & \\
        Memory bank length & \multicolumn{3}{c}{10} \\
        Prompt       &  \multicolumn{3}{c}{``Question: \{\} Short Answer:''}  \\
        \bottomrule
    \end{tabular}
    }
    \label{tab:hyper-vqa}
\end{table}

\begin{table}[h]
\centering
    \caption{Hyperparameters of different datasets on the video captioning task.}
    \vspace{-0.1in}
    \resizebox{\linewidth}{!}{
    \renewcommand{\arraystretch}{1.2}
    \begin{tabular}{@{}l|ccc@{}}
        \toprule
        \textbf{Dataset} & \textbf{MSRVTT} & \textbf{MSVD} & \textbf{YouCook2} \\ \midrule
        LLM & \multicolumn{3}{c}{Vicuna-7B} \\
        Epochs & \multicolumn{3}{c}{10} \\
        Learning rate & 1e-5 & 1e-5 & 1e-4 \\
        Batch size   & \multicolumn{3}{c}{128}  \\
        AdamW \(\beta\)  & \multicolumn{3}{c}{(0.9, 0.999)} \\
        Weight decay & \multicolumn{3}{c}{0.05} \\
        Beam size & \multicolumn{3}{c}{5} \\
        Image resolution & \multicolumn{3}{c}{224} \\
        Frame length & \multicolumn{3}{c}{80} \\
        Memory bank length & \multicolumn{3}{c}{40} \\
        Prompt       &  \multicolumn{3}{c}{``what does the video describe?''}  \\
        \bottomrule
    \end{tabular}
    }
    \label{tab:hyper-caption}
\end{table}

%% file: main.bbl
\begin{thebibliography}{10}

\bibitem{radford2018improving}
Alec Radford, Karthik Narasimhan, Tim Salimans, Ilya Sutskever, et~al.
\newblock Improving language understanding by generative pre-training.
\newblock {\em OpenAI}, 2018.

\bibitem{radford2019language}
Alec Radford, Jeffrey Wu, Rewon Child, David Luan, Dario Amodei, Ilya Sutskever, et~al.
\newblock Language models are unsupervised multitask learners.
\newblock {\em OpenAI blog}, 1(8):9, 2019.

\bibitem{brown2020language}
Tom Brown, Benjamin Mann, Nick Ryder, Melanie Subbiah, Jared~D Kaplan, Prafulla Dhariwal, Arvind Neelakantan, Pranav Shyam, Girish Sastry, Amanda Askell, et~al.
\newblock Language models are few-shot learners.
\newblock {\em Advances in neural information processing systems}, 33:1877--1901, 2020.

\bibitem{chatgpt}
OpenAI.
\newblock Chatgpt.
\newblock {\em https://openai.com/blog/chatgpt}, 2023.

\bibitem{touvron2023llama1}
Hugo Touvron, Thibaut Lavril, Gautier Izacard, Xavier Martinet, Marie-Anne Lachaux, Timoth{\'e}e Lacroix, Baptiste Rozi{\`e}re, Naman Goyal, Eric Hambro, Faisal Azhar, et~al.
\newblock Llama: Open and efficient foundation language models.
\newblock {\em arXiv preprint arXiv:2302.13971}, 2023.

\bibitem{touvron2023llama2}
Hugo Touvron, Louis Martin, Kevin Stone, Peter Albert, Amjad Almahairi, Yasmine Babaei, Nikolay Bashlykov, Soumya Batra, Prajjwal Bhargava, Shruti Bhosale, et~al.
\newblock Llama 2: Open foundation and fine-tuned chat models.
\newblock {\em arXiv preprint arXiv:2307.09288}, 2023.

\bibitem{li2023blip}
Junnan Li, Dongxu Li, Silvio Savarese, and Steven Hoi.
\newblock Blip-2: Bootstrapping language-image pre-training with frozen image encoders and large language models.
\newblock {\em arXiv preprint arXiv:2301.12597}, 2023.

\bibitem{liu2023visual}
Haotian Liu, Chunyuan Li, Qingyang Wu, and Yong~Jae Lee.
\newblock Visual instruction tuning.
\newblock {\em arXiv preprint arXiv:2304.08485}, 2023.

\bibitem{instructblip}
Wenliang Dai, Junnan Li, Dongxu Li, Anthony Meng~Huat Tiong, Junqi Zhao, Weisheng Wang, Boyang Li, Pascale Fung, and Steven Hoi.
\newblock Instructblip: Towards general-purpose vision-language models with instruction tuning, 2023.

\bibitem{zhu2023minigpt}
Deyao Zhu, Jun Chen, Xiaoqian Shen, Xiang Li, and Mohamed Elhoseiny.
\newblock Minigpt-4: Enhancing vision-language understanding with advanced large language models.
\newblock {\em arXiv preprint arXiv:2304.10592}, 2023.

\bibitem{ye2023mplug}
Qinghao Ye, Haiyang Xu, Guohai Xu, Jiabo Ye, Ming Yan, Yiyang Zhou, Junyang Wang, Anwen Hu, Pengcheng Shi, Yaya Shi, et~al.
\newblock mplug-owl: Modularization empowers large language models with multimodality.
\newblock {\em arXiv preprint arXiv:2304.14178}, 2023.

\bibitem{zhang2023video}
Hang Zhang, Xin Li, and Lidong Bing.
\newblock Video-llama: An instruction-tuned audio-visual language model for video understanding.
\newblock {\em arXiv preprint arXiv:2306.02858}, 2023.

\bibitem{2023videochat}
Kunchang Li, Yinan He, Yi~Wang, Yizhuo Li, Wenhai Wang, Ping Luo, Yali Wang, Limin Wang, and Yu~Qiao.
\newblock Videochat: Chat-centric video understanding.
\newblock {\em arXiv preprint arXiv:2305.06355}, 2023.

\bibitem{chen2023videollm}
Guo Chen, Yin-Dong Zheng, Jiahao Wang, Jilan Xu, Yifei Huang, Junting Pan, Yi~Wang, Yali Wang, Yu~Qiao, Tong Lu, et~al.
\newblock Videollm: Modeling video sequence with large language models.
\newblock {\em arXiv preprint arXiv:2305.13292}, 2023.

\bibitem{wang2023visionllm}
Wenhai Wang, Zhe Chen, Xiaokang Chen, Jiannan Wu, Xizhou Zhu, Gang Zeng, Ping Luo, Tong Lu, Jie Zhou, Yu~Qiao, et~al.
\newblock Visionllm: Large language model is also an open-ended decoder for vision-centric tasks.
\newblock {\em arXiv preprint arXiv:2305.11175}, 2023.

\bibitem{chen2023minigpt}
Jun Chen, Deyao Zhu, Xiaoqian Shen, Xiang Li, Zechu Liu, Pengchuan Zhang, Raghuraman Krishnamoorthi, Vikas Chandra, Yunyang Xiong, and Mohamed Elhoseiny.
\newblock Minigpt-v2: Large language model as a unified interface for vision-language multi-task learning.
\newblock {\em arXiv preprint arXiv:2310.09478}, 2023.

\bibitem{wang2022omnivl}
Junke Wang, Dongdong Chen, Zuxuan Wu, Chong Luo, Luowei Zhou, Yucheng Zhao, Yujia Xie, Ce~Liu, Yu-Gang Jiang, and Lu~Yuan.
\newblock Omnivl: One foundation model for image-language and video-language tasks.
\newblock In {\em NeurIPS}, 2022.

\bibitem{wang2022efficient}
Junke Wang, Xitong Yang, Hengduo Li, Li~Liu, Zuxuan Wu, and Yu-Gang Jiang.
\newblock Efficient video transformers with spatial-temporal token selection.
\newblock In {\em ECCV}, 2022.

\bibitem{wang2023omnitracker}
Junke Wang, Dongdong Chen, Zuxuan Wu, Chong Luo, Xiyang Dai, Lu~Yuan, and Yu-Gang Jiang.
\newblock Omnitracker: Unifying object tracking by tracking-with-detection.
\newblock {\em arXiv preprint arXiv:2303.12079}, 2023.

\bibitem{wang2024omnivid}
Junke Wang, Dongdong Chen, Chong Luo, Bo~He, Lu~Yuan, Zuxuan Wu, and Yu-Gang Jiang.
\newblock Omnivid: A generative framework for universal video understanding.
\newblock In {\em CVPR}, 2024.

\bibitem{maaz2023video}
Muhammad Maaz, Hanoona Rasheed, Salman Khan, and Fahad~Shahbaz Khan.
\newblock Video-chatgpt: Towards detailed video understanding via large vision and language models.
\newblock {\em arXiv preprint arXiv:2306.05424}, 2023.

\bibitem{alayrac2022flamingo}
Jean-Baptiste Alayrac, Jeff Donahue, Pauline Luc, Antoine Miech, Iain Barr, Yana Hasson, Karel Lenc, Arthur Mensch, Katherine Millican, Malcolm Reynolds, et~al.
\newblock Flamingo: a visual language model for few-shot learning.
\newblock {\em Advances in Neural Information Processing Systems}, 35:23716--23736, 2022.

\bibitem{hu2021lora}
Edward~J Hu, Yelong Shen, Phillip Wallis, Zeyuan Allen-Zhu, Yuanzhi Li, Shean Wang, Lu~Wang, and Weizhu Chen.
\newblock Lora: Low-rank adaptation of large language models.
\newblock {\em arXiv preprint arXiv:2106.09685}, 2021.

\bibitem{bolya2022token}
Daniel Bolya, Cheng-Yang Fu, Xiaoliang Dai, Peizhao Zhang, Christoph Feichtenhofer, and Judy Hoffman.
\newblock Token merging: Your vit but faster.
\newblock {\em arXiv preprint arXiv:2210.09461}, 2022.

\bibitem{ren2023testa}
Shuhuai Ren, Sishuo Chen, Shicheng Li, Xu~Sun, and Lu~Hou.
\newblock Testa: Temporal-spatial token aggregation for long-form video-language understanding.
\newblock {\em EMNLP}, 2023.

\bibitem{song2023moviechat}
Enxin Song, Wenhao Chai, Guanhong Wang, Yucheng Zhang, Haoyang Zhou, Feiyang Wu, Xun Guo, Tian Ye, Yan Lu, Jenq-Neng Hwang, et~al.
\newblock Moviechat: From dense token to sparse memory for long video understanding.
\newblock {\em CVPR}, 2024.

\bibitem{jin2023chat}
Peng Jin, Ryuichi Takanobu, Caiwan Zhang, Xiaochun Cao, and Li~Yuan.
\newblock Chat-univi: Unified visual representation empowers large language models with image and video understanding.
\newblock {\em CVPR}, 2024.

\bibitem{donahue2015long}
Jeffrey Donahue, Lisa Anne~Hendricks, Sergio Guadarrama, Marcus Rohrbach, Subhashini Venugopalan, Kate Saenko, and Trevor Darrell.
\newblock Long-term recurrent convolutional networks for visual recognition and description.
\newblock In {\em Proceedings of the IEEE conference on computer vision and pattern recognition}, pages 2625--2634, 2015.

\bibitem{yue2015beyond}
Joe Yue-Hei~Ng, Matthew Hausknecht, Sudheendra Vijayanarasimhan, Oriol Vinyals, Rajat Monga, and George Toderici.
\newblock Beyond short snippets: Deep networks for video classification.
\newblock In {\em Proceedings of the IEEE conference on computer vision and pattern recognition}, pages 4694--4702, 2015.

\bibitem{girdhar2017actionvlad}
Rohit Girdhar, Deva Ramanan, Abhinav Gupta, Josef Sivic, and Bryan Russell.
\newblock Actionvlad: Learning spatio-temporal aggregation for action classification.
\newblock In {\em Proceedings of the IEEE conference on computer vision and pattern recognition}, pages 971--980, 2017.

\bibitem{hussein2019timeception}
Noureldien Hussein, Efstratios Gavves, and Arnold~WM Smeulders.
\newblock Timeception for complex action recognition.
\newblock In {\em Proceedings of the IEEE/CVF Conference on Computer Vision and Pattern Recognition}, pages 254--263, 2019.

\bibitem{wu2021towards}
Chao-Yuan Wu and Philipp Krahenbuhl.
\newblock Towards long-form video understanding.
\newblock In {\em Proceedings of the IEEE/CVF Conference on Computer Vision and Pattern Recognition}, pages 1884--1894, 2021.

\bibitem{korbar2019scsampler}
Bruno Korbar, Du~Tran, and Lorenzo Torresani.
\newblock Scsampler: Sampling salient clips from video for efficient action recognition.
\newblock In {\em Proceedings of the IEEE/CVF International Conference on Computer Vision}, pages 6232--6242, 2019.

\bibitem{wu2019adaframe}
Zuxuan Wu, Caiming Xiong, Chih-Yao Ma, Richard Socher, and Larry~S Davis.
\newblock Adaframe: Adaptive frame selection for fast video recognition.
\newblock In {\em Proceedings of the IEEE/CVF Conference on Computer Vision and Pattern Recognition}, pages 1278--1287, 2019.

\bibitem{islam2022long}
Md~Mohaiminul Islam and Gedas Bertasius.
\newblock Long movie clip classification with state-space video models.
\newblock In {\em European Conference on Computer Vision}, pages 87--104. Springer, 2022.

\bibitem{wang2023selective}
Jue Wang, Wentao Zhu, Pichao Wang, Xiang Yu, Linda Liu, Mohamed Omar, and Raffay Hamid.
\newblock Selective structured state-spaces for long-form video understanding.
\newblock In {\em Proceedings of the IEEE/CVF Conference on Computer Vision and Pattern Recognition}, pages 6387--6397, 2023.

\bibitem{gu2021efficiently}
Albert Gu, Karan Goel, and Christopher R{\'e}.
\newblock Efficiently modeling long sequences with structured state spaces.
\newblock {\em arXiv preprint arXiv:2111.00396}, 2021.

\bibitem{chen2020memory}
Yihong Chen, Yue Cao, Han Hu, and Liwei Wang.
\newblock Memory enhanced global-local aggregation for video object detection.
\newblock In {\em Proceedings of the IEEE/CVF conference on computer vision and pattern recognition}, pages 10337--10346, 2020.

\bibitem{lee2018memory}
Sangho Lee, Jinyoung Sung, Youngjae Yu, and Gunhee Kim.
\newblock A memory network approach for story-based temporal summarization of 360 videos.
\newblock In {\em Proceedings of the IEEE conference on computer vision and pattern recognition}, pages 1410--1419, 2018.

\bibitem{lee2021video}
Sangmin Lee, Hak~Gu Kim, Dae~Hwi Choi, Hyung-Il Kim, and Yong~Man Ro.
\newblock Video prediction recalling long-term motion context via memory alignment learning.
\newblock In {\em Proceedings of the IEEE/CVF Conference on Computer Vision and Pattern Recognition}, pages 3054--3063, 2021.

\bibitem{wu2022memvit}
Chao-Yuan Wu, Yanghao Li, Karttikeya Mangalam, Haoqi Fan, Bo~Xiong, Jitendra Malik, and Christoph Feichtenhofer.
\newblock Memvit: Memory-augmented multiscale vision transformer for efficient long-term video recognition.
\newblock In {\em Proceedings of the IEEE/CVF Conference on Computer Vision and Pattern Recognition}, pages 13587--13597, 2022.

\bibitem{feichtenhofer2016convolutional}
Christoph Feichtenhofer, Axel Pinz, and Andrew Zisserman.
\newblock Convolutional two-stream network fusion for video action recognition.
\newblock In {\em Proceedings of the IEEE conference on computer vision and pattern recognition}, pages 1933--1941, 2016.

\bibitem{feichtenhofer2019slowfast}
Christoph Feichtenhofer, Haoqi Fan, Jitendra Malik, and Kaiming He.
\newblock Slowfast networks for video recognition.
\newblock In {\em Proceedings of the IEEE/CVF international conference on computer vision}, pages 6202--6211, 2019.

\bibitem{he2020gta}
Bo~He, Xitong Yang, Zuxuan Wu, Hao Chen, Ser-Nam Lim, and Abhinav Shrivastava.
\newblock Gta: Global temporal attention for video action understanding.
\newblock {\em British Machine Vision Conference (BMVC)}, 2021.

\bibitem{he2022asm}
Bo~He, Xitong Yang, Le~Kang, Zhiyu Cheng, Xin Zhou, and Abhinav Shrivastava.
\newblock Asm-loc: action-aware segment modeling for weakly-supervised temporal action localization.
\newblock In {\em Proceedings of the IEEE/CVF Conference on Computer Vision and Pattern Recognition}, pages 13925--13935, 2022.

\bibitem{he2023align}
Bo~He, Jun Wang, Jielin Qiu, Trung Bui, Abhinav Shrivastava, and Zhaowen Wang.
\newblock Align and attend: Multimodal summarization with dual contrastive losses.
\newblock In {\em Proceedings of the IEEE/CVF Conference on Computer Vision and Pattern Recognition}, pages 14867--14878, 2023.

\bibitem{bertasius2021space}
Gedas Bertasius, Heng Wang, and Lorenzo Torresani.
\newblock Is space-time attention all you need for video understanding?
\newblock In {\em ICML}, volume~2, page~4, 2021.

\bibitem{fan2021multiscale}
Haoqi Fan, Bo~Xiong, Karttikeya Mangalam, Yanghao Li, Zhicheng Yan, Jitendra Malik, and Christoph Feichtenhofer.
\newblock Multiscale vision transformers.
\newblock In {\em Proceedings of the IEEE/CVF international conference on computer vision}, pages 6824--6835, 2021.

\bibitem{tong2022videomae}
Zhan Tong, Yibing Song, Jue Wang, and Limin Wang.
\newblock Videomae: Masked autoencoders are data-efficient learners for self-supervised video pre-training.
\newblock {\em Advances in neural information processing systems}, 35:10078--10093, 2022.

\bibitem{meng2022adavit}
Lingchen Meng, Hengduo Li, Bor-Chun Chen, Shiyi Lan, Zuxuan Wu, Yu-Gang Jiang, and Ser-Nam Lim.
\newblock Adavit: Adaptive vision transformers for efficient image recognition.
\newblock In {\em Proceedings of the IEEE/CVF Conference on Computer Vision and Pattern Recognition}, pages 12309--12318, 2022.

\bibitem{yin2022vit}
Hongxu Yin, Arash Vahdat, Jose~M Alvarez, Arun Mallya, Jan Kautz, and Pavlo Molchanov.
\newblock A-vit: Adaptive tokens for efficient vision transformer.
\newblock In {\em Proceedings of the IEEE/CVF Conference on Computer Vision and Pattern Recognition}, pages 10809--10818, 2022.

\bibitem{rao2021dynamicvit}
Yongming Rao, Wenliang Zhao, Benlin Liu, Jiwen Lu, Jie Zhou, and Cho-Jui Hsieh.
\newblock Dynamicvit: Efficient vision transformers with dynamic token sparsification.
\newblock {\em Advances in neural information processing systems}, 34:13937--13949, 2021.

\bibitem{choromanski2020rethinking}
Krzysztof Choromanski, Valerii Likhosherstov, David Dohan, Xingyou Song, Andreea Gane, Tamas Sarlos, Peter Hawkins, Jared Davis, Afroz Mohiuddin, Lukasz Kaiser, et~al.
\newblock Rethinking attention with performers.
\newblock {\em arXiv preprint arXiv:2009.14794}, 2020.

\bibitem{patrick2021keeping}
Mandela Patrick, Dylan Campbell, Yuki Asano, Ishan Misra, Florian Metze, Christoph Feichtenhofer, Andrea Vedaldi, and Joao~F Henriques.
\newblock Keeping your eye on the ball: Trajectory attention in video transformers.
\newblock {\em Advances in neural information processing systems}, 34:12493--12506, 2021.

\bibitem{sun2019videobert}
Chen Sun, Austin Myers, Carl Vondrick, Kevin Murphy, and Cordelia Schmid.
\newblock Videobert: A joint model for video and language representation learning.
\newblock In {\em Proceedings of the IEEE/CVF international conference on computer vision}, pages 7464--7473, 2019.

\bibitem{kuehne2014language}
Hilde Kuehne, Ali Arslan, and Thomas Serre.
\newblock The language of actions: Recovering the syntax and semantics of goal-directed human activities.
\newblock In {\em Proceedings of the IEEE conference on computer vision and pattern recognition}, pages 780--787, 2014.

\bibitem{tang2019coin}
Yansong Tang, Dajun Ding, Yongming Rao, Yu~Zheng, Danyang Zhang, Lili Zhao, Jiwen Lu, and Jie Zhou.
\newblock Coin: A large-scale dataset for comprehensive instructional video analysis.
\newblock In {\em Proceedings of the IEEE/CVF Conference on Computer Vision and Pattern Recognition}, pages 1207--1216, 2019.

\bibitem{wang2018temporal}
Limin Wang, Yuanjun Xiong, Zhe Wang, Yu~Qiao, Dahua Lin, Xiaoou Tang, and Luc Van~Gool.
\newblock Temporal segment networks for action recognition in videos.
\newblock {\em IEEE transactions on pattern analysis and machine intelligence}, 41(11):2740--2755, 2018.

\bibitem{hussein2019videograph}
Noureldien Hussein, Efstratios Gavves, and Arnold~WM Smeulders.
\newblock Videograph: Recognizing minutes-long human activities in videos.
\newblock {\em arXiv preprint arXiv:1905.05143}, 2019.

\bibitem{zhou2021graph}
Jiaming Zhou, Kun-Yu Lin, Haoxin Li, and Wei-Shi Zheng.
\newblock Graph-based high-order relation modeling for long-term action recognition.
\newblock In {\em Proceedings of the IEEE/CVF Conference on Computer Vision and Pattern Recognition}, pages 8984--8993, 2021.

\bibitem{lin2022learning}
Xudong Lin, Fabio Petroni, Gedas Bertasius, Marcus Rohrbach, Shih-Fu Chang, and Lorenzo Torresani.
\newblock Learning to recognize procedural activities with distant supervision.
\newblock In {\em Proceedings of the IEEE/CVF Conference on Computer Vision and Pattern Recognition}, pages 13853--13863, 2022.

\bibitem{xu2017video}
Dejing Xu, Zhou Zhao, Jun Xiao, Fei Wu, Hanwang Zhang, Xiangnan He, and Yueting Zhuang.
\newblock Video question answering via gradually refined attention over appearance and motion.
\newblock In {\em Proceedings of the 25th ACM international conference on Multimedia}, pages 1645--1653, 2017.

\bibitem{yu2019activitynet}
Zhou Yu, Dejing Xu, Jun Yu, Ting Yu, Zhou Zhao, Yueting Zhuang, and Dacheng Tao.
\newblock Activitynet-qa: A dataset for understanding complex web videos via question answering.
\newblock In {\em Proceedings of the AAAI Conference on Artificial Intelligence}, volume~33, pages 9127--9134, 2019.

\bibitem{banerjee2005meteor}
Satanjeev Banerjee and Alon Lavie.
\newblock Meteor: An automatic metric for mt evaluation with improved correlation with human judgments.
\newblock In {\em Proceedings of the acl workshop on intrinsic and extrinsic evaluation measures for machine translation and/or summarization}, pages 65--72, 2005.

\bibitem{vedantam2015cider}
Ramakrishna Vedantam, C~Lawrence~Zitnick, and Devi Parikh.
\newblock Cider: Consensus-based image description evaluation.
\newblock In {\em Proceedings of the IEEE conference on computer vision and pattern recognition}, pages 4566--4575, 2015.

\bibitem{xu2016msr}
Jun Xu, Tao Mei, Ting Yao, and Yong Rui.
\newblock Msr-vtt: A large video description dataset for bridging video and language.
\newblock In {\em Proceedings of the IEEE conference on computer vision and pattern recognition}, pages 5288--5296, 2016.

\bibitem{chen2011collecting}
David~L Chen and William~B Dolan.
\newblock Collecting highly parallel data for paraphrase evaluation.
\newblock In {\em Proceedings of the 49th Annual Meeting of the Association for Computational Linguistics: Human Language Technologies-Volume 1}, pages 190--200. Association for Computational Linguistics, 2011.

\bibitem{zhou2018towards}
Luowei Zhou, Chenliang Xu, and Jason Corso.
\newblock Towards automatic learning of procedures from web instructional videos.
\newblock In {\em Proceedings of the AAAI Conference on Artificial Intelligence}, volume~32, 2018.

\bibitem{damen2020epic}
Dima Damen, Hazel Doughty, Giovanni~Maria Farinella, Sanja Fidler, Antonino Furnari, Evangelos Kazakos, Davide Moltisanti, Jonathan Munro, Toby Perrett, Will Price, et~al.
\newblock The epic-kitchens dataset: Collection, challenges and baselines.
\newblock {\em IEEE Transactions on Pattern Analysis and Machine Intelligence}, 43(11):4125--4141, 2020.

\bibitem{zhong2023anticipative}
Zeyun Zhong, David Schneider, Michael Voit, Rainer Stiefelhagen, and J{\"u}rgen Beyerer.
\newblock Anticipative feature fusion transformer for multi-modal action anticipation.
\newblock In {\em Proceedings of the IEEE/CVF Winter Conference on Applications of Computer Vision}, pages 6068--6077, 2023.

\bibitem{dosovitskiy2020image}
Alexey Dosovitskiy, Lucas Beyer, Alexander Kolesnikov, Dirk Weissenborn, Xiaohua Zhai, Thomas Unterthiner, Mostafa Dehghani, Matthias Minderer, Georg Heigold, Sylvain Gelly, et~al.
\newblock An image is worth 16x16 words: Transformers for image recognition at scale.
\newblock {\em arXiv preprint arXiv:2010.11929}, 2020.

\bibitem{fang2023eva}
Yuxin Fang, Wen Wang, Binhui Xie, Quan Sun, Ledell Wu, Xinggang Wang, Tiejun Huang, Xinlong Wang, and Yue Cao.
\newblock Eva: Exploring the limits of masked visual representation learning at scale.
\newblock In {\em Proceedings of the IEEE/CVF Conference on Computer Vision and Pattern Recognition}, pages 19358--19369, 2023.

\bibitem{vicuna2023}
Wei-Lin Chiang, Zhuohan Li, Zi~Lin, Ying Sheng, Zhanghao Wu, Hao Zhang, Lianmin Zheng, Siyuan Zhuang, Yonghao Zhuang, Joseph~E. Gonzalez, Ion Stoica, and Eric~P. Xing.
\newblock Vicuna: An open-source chatbot impressing gpt-4 with 90\%* chatgpt quality, March 2023.

\bibitem{yang2021just}
Antoine Yang, Antoine Miech, Josef Sivic, Ivan Laptev, and Cordelia Schmid.
\newblock Just ask: Learning to answer questions from millions of narrated videos.
\newblock In {\em Proceedings of the IEEE/CVF international conference on computer vision}, pages 1686--1697, 2021.

\bibitem{yang2022zero}
Antoine Yang, Antoine Miech, Josef Sivic, Ivan Laptev, and Cordelia Schmid.
\newblock Zero-shot video question answering via frozen bidirectional language models.
\newblock {\em Advances in Neural Information Processing Systems}, 35:124--141, 2022.

\bibitem{lei2022revealing}
Jie Lei, Tamara~L Berg, and Mohit Bansal.
\newblock Revealing single frame bias for video-and-language learning.
\newblock {\em arXiv preprint arXiv:2206.03428}, 2022.

\bibitem{fu2023empirical}
Tsu-Jui Fu, Linjie Li, Zhe Gan, Kevin Lin, William~Yang Wang, Lijuan Wang, and Zicheng Liu.
\newblock An empirical study of end-to-end video-language transformers with masked visual modeling.
\newblock In {\em Proceedings of the IEEE/CVF Conference on Computer Vision and Pattern Recognition}, pages 22898--22909, 2023.

\bibitem{wang2022git}
Jianfeng Wang, Zhengyuan Yang, Xiaowei Hu, Linjie Li, Kevin Lin, Zhe Gan, Zicheng Liu, Ce~Liu, and Lijuan Wang.
\newblock Git: A generative image-to-text transformer for vision and language.
\newblock {\em arXiv preprint arXiv:2205.14100}, 2022.

\bibitem{xu2023mplug}
Haiyang Xu, Qinghao Ye, Ming Yan, Yaya Shi, Jiabo Ye, Yuanhong Xu, Chenliang Li, Bin Bi, Qi~Qian, Wei Wang, et~al.
\newblock mplug-2: A modularized multi-modal foundation model across text, image and video.
\newblock {\em arXiv preprint arXiv:2302.00402}, 2023.

\bibitem{li2023unmasked}
Kunchang Li, Yali Wang, Yizhuo Li, Yi~Wang, Yinan He, Limin Wang, and Yu~Qiao.
\newblock Unmasked teacher: Towards training-efficient video foundation models, 2023.

\bibitem{yan2022video}
Shen Yan, Tao Zhu, Zirui Wang, Yuan Cao, Mi~Zhang, Soham Ghosh, Yonghui Wu, and Jiahui Yu.
\newblock Video-text modeling with zero-shot transfer from contrastive captioners.
\newblock {\em arXiv preprint arXiv:2212.04979}, 2022.

\bibitem{luo2020univl}
Huaishao Luo, Lei Ji, Botian Shi, Haoyang Huang, Nan Duan, Tianrui Li, Jason Li, Taroon Bharti, and Ming Zhou.
\newblock Univl: A unified video and language pre-training model for multimodal understanding and generation.
\newblock {\em arXiv preprint arXiv:2002.06353}, 2020.

\bibitem{lin2022swinbert}
Kevin Lin, Linjie Li, Chung-Ching Lin, Faisal Ahmed, Zhe Gan, Zicheng Liu, Yumao Lu, and Lijuan Wang.
\newblock Swinbert: End-to-end transformers with sparse attention for video captioning.
\newblock In {\em Proceedings of the IEEE/CVF Conference on Computer Vision and Pattern Recognition}, pages 17949--17958, 2022.

\bibitem{miech2019howto100m}
Antoine Miech, Dimitri Zhukov, Jean-Baptiste Alayrac, Makarand Tapaswi, Ivan Laptev, and Josef Sivic.
\newblock Howto100m: Learning a text-video embedding by watching hundred million narrated video clips.
\newblock In {\em Proceedings of the IEEE/CVF international conference on computer vision}, pages 2630--2640, 2019.

\bibitem{nagrani2022learning}
Arsha Nagrani, Paul~Hongsuck Seo, Bryan Seybold, Anja Hauth, Santiago Manen, Chen Sun, and Cordelia Schmid.
\newblock Learning audio-video modalities from image captions.
\newblock In {\em European Conference on Computer Vision}, pages 407--426. Springer, 2022.

\bibitem{chung2022scaling}
Hyung~Won Chung, Le~Hou, Shayne Longpre, Barret Zoph, Yi~Tay, William Fedus, Yunxuan Li, Xuezhi Wang, Mostafa Dehghani, Siddhartha Brahma, et~al.
\newblock Scaling instruction-finetuned language models.
\newblock {\em arXiv preprint arXiv:2210.11416}, 2022.

\bibitem{wu2019long}
Chao-Yuan Wu, Christoph Feichtenhofer, Haoqi Fan, Kaiming He, Philipp Krahenbuhl, and Ross Girshick.
\newblock Long-term feature banks for detailed video understanding.
\newblock In {\em Proceedings of the IEEE/CVF conference on computer vision and pattern recognition}, pages 284--293, 2019.

\bibitem{gong2019memorizing}
Dong Gong, Lingqiao Liu, Vuong Le, Budhaditya Saha, Moussa~Reda Mansour, Svetha Venkatesh, and Anton van~den Hengel.
\newblock Memorizing normality to detect anomaly: Memory-augmented deep autoencoder for unsupervised anomaly detection.
\newblock In {\em Proceedings of the IEEE/CVF international conference on computer vision}, pages 1705--1714, 2019.

\bibitem{li-etal-2023-lavis}
Dongxu Li, Junnan Li, Hung Le, Guangsen Wang, Silvio Savarese, and Steven~C.H. Hoi.
\newblock {LAVIS}: A one-stop library for language-vision intelligence.
\newblock In {\em Proceedings of the 61st Annual Meeting of the Association for Computational Linguistics (Volume 3: System Demonstrations)}, pages 31--41, Toronto, Canada, July 2023. Association for Computational Linguistics.

\end{thebibliography}
